\pdfoutput=1

\documentclass[11pt]{article}

\usepackage[preprint]{acl}

\usepackage{times}
\usepackage{latexsym}
\usepackage{listings}
\usepackage[T1]{fontenc}

\usepackage[utf8]{inputenc}

\usepackage{microtype}

\usepackage{inconsolata}

\usepackage{graphicx}
\usepackage{amsmath}
\usepackage{tabularx}
\usepackage{booktabs}
\usepackage{arydshln}
\usepackage{multirow}
\usepackage{multicol}
\usepackage{colortbl}
\usepackage{quoting}

\makeatletter
\def\adl@drawiv#1#2#3{%
        \xleaders#3{#2.3\@tempdimb #1{1}#2.3\@tempdimb}%
                #2\z@ plus1fil minus1fil\relax
        }
\newcommand{\cdashlinelr}[1]{%
  \noalign{\vskip\aboverulesep
           \global\let\@dashdrawstore\adl@draw
           \global\let\adl@draw\adl@drawiv}
  \cdashline{#1}
  \noalign{\global\let\adl@draw\@dashdrawstore
           \vskip\belowrulesep}}
\makeatother

\definecolor{myred}{RGB}{200, 20, 20}
\definecolor{mygreen}{RGB}{15, 140, 0}
\newcommand\red[1]{\textcolor{myred}{#1}}
\newcommand\green[1]{\textcolor{mygreen}{#1}}

\title{Can LLM Generate Culturally Relevant Commonsense QA Data? \\Case Study in Indonesian and Sundanese}

\author{Rifki Afina Putri$^{1}$, Faiz Ghifari Haznitrama$^{1}$, Dea Adhista$^{2}$, Alice Oh$^{1}$ \\
        $^1$KAIST, Republic of Korea $^2$Independent Researcher \\
        \texttt{\{rifkiaputri,haznitrama\}@kaist.ac.kr}, \texttt{deadhista@gmail.com}, \texttt{alice.oh@kaist.edu}
        }

\begin{document}
\maketitle
\begin{abstract}
Large Language Models (LLMs) are increasingly being used to generate synthetic data for training and evaluating models. However, it is unclear whether they can generate a good quality of question answering (QA) dataset that incorporates knowledge and cultural nuance embedded in a language, especially for low-resource languages. In this study, we investigate the effectiveness of using LLMs in generating culturally relevant commonsense QA datasets for Indonesian and Sundanese languages. To do so, we create datasets for these languages using various methods involving both LLMs and human annotators, resulting in $\sim$4.5K questions per language ($\sim$9K in total), making our dataset the largest of its kind. Our experiments show that automatic data adaptation from an existing English dataset is less effective for Sundanese. Interestingly, using the direct generation method on the target language, GPT-4 Turbo can generate questions with adequate general knowledge in both languages, albeit not as culturally `deep' as humans. We also observe a higher occurrence of fluency errors in the Sundanese dataset, highlighting the discrepancy between medium- and lower-resource languages.\footnote{All datasets and codes in this work are available at \url{https://github.com/rifkiaputri/id-csqa}.}
\end{abstract}

\section{Introduction}
The development of Large Language Models (LLMs) is significantly impacting NLP, leading to an increasing trend in the automated generation of datasets, particularly for question answering (QA) tasks. However, a major challenge arises with underrepresented languages like Indonesian and Sundanese due to the need for cultural context. For the generated data to be fully useful, it must not only be linguistically accurate, but it also needs to reflect the cultural nuances, historical references, and social norms. It is not yet clear whether current LLMs can create QA data that adequately includes the cultural nuances specific to certain languages.

Another common method for constructing non-English datasets is by using machine translation. Although more scalable, this method cannot be straightforwardly applied due to the contextual irrelevancy of the data, primarily due to geographical differences (which could also influence cultural differences). For example, many English CommonsenseQA questions \cite{talmor-etal-2019-commonsenseqa} include concepts such as \textit{snow} or any \textit{winter sports}, which are irrelevant in Indonesia due to its tropical climate with only two seasons. This dataset also often includes English-centric names and locations, primarily limited to the US, which are not considered ``commonsense'' for Indonesians.

\begin{figure*}[t!]
    \begin{center}
        \centerline{\includegraphics[width=\textwidth]{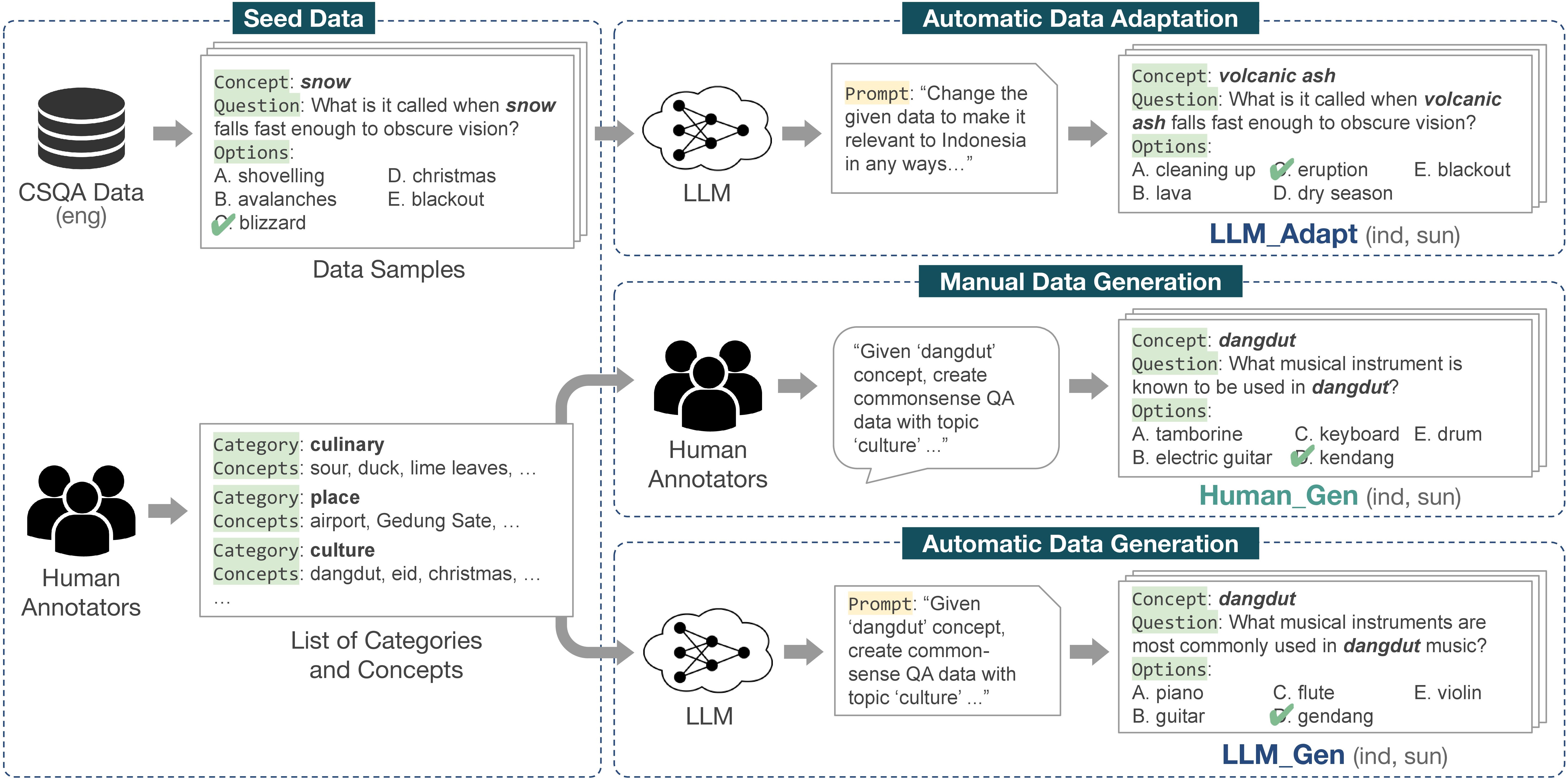}}
        \vspace{-4pt}
        \caption{Our dataset generation methods. The examples of \textsc{LLM\_Adapt}, \textsc{Human\_Gen}, and \textsc{LLM\_Gen} datasets are shown in English for clarity. The original versions of these datasets are in Indonesian and Sundanese.}
        \label{fig:data_gen_method}
    \end{center}
\end{figure*}

Therefore, in this study, we aim to investigate how well current LLMs can adapt and generate a commonsense QA dataset that is both linguistically accurate and culturally relevant to Indonesia. We focus on Indonesian, the lingua franca of Indonesia, and Sundanese, one of the regional languages in Indonesia with 34 million speakers, yet still considered low-resource \cite{aji-etal-2022-one}. To address dataset scarcity, especially in Sundanese, we also manually constructed datasets by involving annotators from various regions in Indonesia, ensuring a representation of diverse cultural perspectives. To sum up, our contributions are as follows:
\begin{itemize}
    \item We create a new Indonesian and Sundanese CommonsenseQA dataset using various methods (Figure \ref{fig:data_gen_method}), including adapting existing English data (\textsc{LLM\_Adapt}) and generating new datasets manually (\textsc{Human\_Gen}) and automatically (\textsc{LLM\_Gen}). The dataset contains $\sim$3K LLM-generated and $\sim$1.5K human-generated question-answer pairs per language, which, to our knowledge, is the largest culturally nuanced commonsense QA dataset in both Indonesian and, particularly, Sundanese.
    \item We perform a comprehensive analysis to assess the effectiveness of LLMs in creating a culturally relevant commonsense QA dataset. We find that adaptation from English data is less effective, especially for Sundanese. However, GPT-4 Turbo can generate questions with some basic local knowledge in Indonesian and Sundanese when provided with human-created categories and concepts.
    \item We evaluate several LLMs on our dataset and find that they perform better on LLM-generated data than on human-generated data, indicating that the former is less challenging, especially for proprietary models like GPT-4. Nevertheless, many open-source LLMs still struggle to answer LLM-generated questions, highlighting significant room for improvement in these datasets.
\end{itemize}

\section{Related Work}
\paragraph{Commonsense Dataset}
Datasets for commonsense reasoning are crucial for developing models that understand and reason about real-world complexities. Datasets like COPA \cite{roemmele2011choice}, X-COPA \cite{ponti-etal-2020-xcopa}, and The Winograd Schema Challenge \cite{levesque2012winograd} evaluate causal reasoning in real-world scenarios. Other datasets like ARC \cite{clark2018think}, OpenBookQA \cite{mihaylov-etal-2018-suit}, and Mcscript \cite{ostermann-etal-2018-mcscript} test commonsense reasoning via QA, but some questions require grade-school science knowledge. CommonsenseQA \cite{talmor-etal-2019-commonsenseqa} presents the type of ``purely'' commonsense QA in the form of multiple-choice questions based on knowledge from ConceptNet \cite{speer2017conceptnet}, built through crowdsourcing. However, the crowdworkers bring their cultural background as part of their common knowledge in building the data, resulting in many questions that are ``commonsense'' only in the Western culture.

\paragraph{Cultural Evaluation Dataset}
Various datasets are used to assess language models' ability to understand cultural nuances specific to a language. FORK \cite{palta-rudinger-2023-fork} explores culinary cultural biases and assumptions of US, Indian, and Chinese customs. IndoMMLU \cite{koto-etal-2023-large} includes questions from Indonesian exams and covers regional cultural topics, such as the Minangkabau or Sundanese cultures. COPAL-ID \cite{wibowo2023copalid} is a COPA-style dataset written by native speakers, thus incorporating more Indonesian cultural nuances compared to X-COPA. COPAL-ID is composed of approximately 500 questions, with a primary emphasis on the cultural aspects of the Jakarta region. In contrast, our dataset is much bigger and covers a broader range of annotators from various regions of Java and Bali, in addition to Jakarta. We also include Sundanese, addressing the gap in this low-resource language.

\section{Background}
\subsection{Commonsense QA: Definition and Scope}
\label{sec:task_scope}
In our dataset, we focus on building questions that probe common or cultural knowledge in daily life within Indonesian and Sundanese contexts. Each data point is a triple of \textit{concept}, \textit{question}, and \textit{options}, with one correct answer. To enhance the relevance of the dataset to local contexts and reduce Western cultural bias, we incorporate both existing question concepts from the English CommonsenseQA \cite{talmor-etal-2019-commonsenseqa} and novel question concepts that we manually created.

Unlike English CommonsenseQA, our dataset includes question category metadata, covering five categories: \textit{culinary}, \textit{place}, \textit{culture}, \textit{history}, and \textit{activity}. Each category has 150 unique question concepts, ensuring broad domain and knowledge coverage.\footnote{More details on how we curate our categories and concepts can be seen in Appendix \ref{appendix:categories_and_concepts}.} Another key difference is that English CommonsenseQA concepts are sourced from a knowledge base, and they divide the concept sets into questions and options. In contrast, our dataset provides only a question concept and its category as input. The options, including the correct answer, are entirely generated by the data creator.

\subsection{Languages in Indonesia}
\label{sec:languages_in_indonesia}
Indonesia is one of the most culturally and linguistically diverse countries, with more than 700 languages spoken across the country \cite{aji-etal-2022-one,Eberhard2021Ethnologue}. Among the many languages, Indonesian is a unifying language used nationally. It utilizes the Latin script and was developed from literary "Classical Malay" \cite{Sneddon2003}, with regional variations. Over 80\% of Standard Malay's vocabulary is similar to Indonesian.

Apart from Indonesian, regional languages like Sundanese are spoken by people of the same ethnicity. It is primarily spoken in West Java and is the second-largest regional language in Indonesia, with 34 million speakers \cite{Eberhard2021Ethnologue}.
Regional languages, including Sundanese, have influenced the formation and development of the Indonesian language. Both languages share similarities, such as their grammatical structure, but also differ significantly in aspects like the number of vowels (i.e., Sundanese has 2 additional vowels: é, eu) and morphological features, including affixes.

We chose to study the national language and one regional language to illustrate the differences in the commonsense QA data generated via LLMs. It is also important to note that despite its large number of speakers, Sundanese has very limited data, and to date, there are no commonsense QA datasets available in Sundanese.

\section{Data Generation Methods}
To investigate whether LLMs can generate culturally relevant commonsense QA data in Indonesian and Sundanese, we build a dataset using various methods with LLMs as data generators. We also employ humans to generate data for comparison.

As illustrated in Figure \ref{fig:data_gen_method}, we apply three dataset generation methods: (1) \textbf{Automatic Data Adaptation} (\textsc{LLM\_Adapt}), where we leverage LLMs to automatically adapt English CommonsenseQA data to our target languages; (2) \textbf{Manual Data Generation} (\textsc{Human\_Gen}), where we ask native-speaker human annotators to manually construct the dataset; and (3) \textbf{Automatic Data Generation} (\textsc{LLM\_Gen}), where we use LLMs to generate data based on the list of categories and concepts used in \textsc{Human\_Gen}. Table \ref{tab:data_stat} shows the statistics of our final dataset, containing a total of \textbf{8,953} QA pairs (4,416 Indonesian and 4,537 Sundanese).

\begin{table}[t]
\centering
\small
\resizebox{\columnwidth}{!}{%
\begin{tabular}{@{}lrrrr@{}}
\toprule
\multicolumn{1}{c}{\multirow{2}{*}{\textbf{\begin{tabular}[c]{@{}c@{}}Dataset\\ Version\end{tabular}}}} &
  \multicolumn{2}{c}{\textbf{Indonesian}} &
  \multicolumn{2}{c}{\textbf{Sundanese}} \\ \cmidrule(lr){2-3} \cmidrule(l){4-5} 
\multicolumn{1}{c}{} &
  \multicolumn{1}{c}{\textbf{Train / Valid / Test}} &
  \multicolumn{1}{c}{\textbf{Total}} &
  \multicolumn{1}{c}{\textbf{Train / Valid / Test}} &
  \multicolumn{1}{c}{\textbf{Total}} \\ \midrule
\textsc{LLM\_Adapt}           & 1,506 / 191 / 158   & \textbf{1,855} & 1,506 / 191 / 158   & \textbf{1,855} \\
\textsc{Human\_Gen}           & 0 / 0 / 1,498       & \textbf{1,498} & 0 / 0 / 1,499       & \textbf{1,499} \\
\textsc{LLM\_Gen}             & 0 / 0 / 1,063       & \textbf{1,063} & 0 / 0 / 1,183       & \textbf{1,183} \\ \midrule
\multicolumn{1}{c}{} & 1,506 / 191 / 2,719 & \textbf{4,416} & 1,506 / 191 / 2,840 & \textbf{4,537} \\ \bottomrule
\end{tabular}%
}
\vspace{-5px}
\caption{Statistics of our generated Indonesian and Sundanese CommonsenseQA dataset. We retained the original English CommonsenseQA splits in \textsc{LLM\_Adapt} to avoid data contamination.}
\label{tab:data_stat}
\end{table}

While all prompts given to the LLM are in English, for the \textsc{LLM\_Adapt} data, we specifically instruct the model to produce responses directly in Indonesian. Subsequently, we translate the resulting Indonesian data into Sundanese. In contrast, for the \textsc{LLM\_Gen} data, we instruct the model to give the output directly in Indonesian and also Sundanese, to closely replicate the data generation process used for \textsc{Human\_Gen}. The generation methods for each dataset are detailed below.

\subsection{Automatic Data Adaptation}
\label{sec:llm_adapt}
We build the first LLM-generated data by adapting English CommonsenseQA dataset \cite{talmor-etal-2019-commonsenseqa} to make it culturally relevant to Indonesian and Sundanese.

\paragraph{Data Selection}
To select the data to be adapted, we first remove data containing offensive keywords and those with duplicate or similar options in different tenses. Unlike English, Indonesian and Sundanese do not change form to indicate tenses (e.g., `passed' and `passing' both translate to \textit{`lulus'}).

Then, we sample the data by assessing three elements: \textit{concept}, \textit{name}, and \textit{location}. Data that are considered irrelevant in at least one of the three elements are selected to be adapted. We take question concepts from the existing CommonsenseQA data and use Stanza \cite{qi2020stanza} and ConceptNet API\footnote{\url{https://github.com/commonsense/conceptnet5/wiki/API}} for name and location extraction.

To determine concept relevance, we utilize GPT-3.5 Turbo with five different prompts to ensure reliable results, asking whether a concept is relevant in Indonesia or West Java.\footnote{All data generation prompts are detailed in Appendix \ref{appendix:data_gen_prompts}.} For example, given the concept `snow', one prompt is, \textit{"Can snow be found in West Java? Answer with only ‘yes’ or ‘no’."} A `yes' means the concept is relevant. Final relevancy is determined by majority voting. Data containing person names and/or locations are regarded as irrelevant by default to save the relevancy classifier API cost, as our preliminary experiment showed all such data are Western-centric. From the total of $\sim$12K QA pairs in English CommonsenseQA, $\sim$2K are selected to be adapted.

\paragraph{QA Pairs Adaptation}
The next step is to transform the selected irrelevant data. First, we prompt GPT-4 Turbo\footnote{Our initial experiment (Appendix \ref{appendix:model_gen_compare}) reveals that GPT-4 Turbo significantly outperforms Merak-v4, an open Indonesian LLM, leading us to select GPT-4 Turbo for our work.} to rephrase the sampled questions and options to align with Indonesian cultures. Subsequently, for data flagged with Western-centric names, we use GPT-3.5 Turbo to replace all person names. The fully adapted data are then translated from Indonesian to Sundanese using Google Translation API. We choose machine translation due to the unreliability of direct adaptation, as the concepts are sourced from English data. Our preliminary run on \texttt{eng} to \texttt{sun} adaptation shows that 90\% of the samples contained errors, with half due to hallucinations. This occurred less frequently in \texttt{eng} to \texttt{ind} adaptation. For example, for the `bald eagle' concept, GPT-4 Turbo generated \textit{`elang Jawa'} (Javan hawk-eagle) in Indonesian, but it generated the non-existent concept \textit{`garuda puspa'} (literally translates to eagle flower) in Sundanese. Although not ideal, translation from the adapted Indonesian QA pairs generally provides better data quality.

\paragraph{Data Filtering}
Finally, we eliminate low-quality data by removing instances where the concepts do not appear in the questions. We also utilize a back-translation method to filter out poor translations of Indonesian (\texttt{ind}) to Sundanese (\texttt{sun}) data. We discard QA pairs if the similarity between the original Indonesian (\texttt{ind}) and the back-translated version (\texttt{ind'}) is less than 0.9. We measure similarity using an embedding-based metric with multilingual MiniLM \cite{reimers-2019-sentence-bert} and LaBSE \cite{feng-etal-2022-language} models, and keep the data if either similarity score is above 0.9.

\subsection{Manual Data Generation}
\label{sec:human_gen}
To build the human-generated dataset, we recruited 12 experienced annotators from diverse regions across Java and Bali. All are native speakers of Indonesian (and Sundanese), with a minimum of 15 years residing in their particular target language regions. Among them, 5 annotators have a degree in linguistics. These criteria ensure they possess both language proficiency and cultural familiarity. As for the data collection process, we have two main phases: (1) creating commonsense question-answer pairs and (2) answering commonsense questions. We also perform quality control to ensure the data quality.\footnote{More details on annotators' demographic information and annotation guidelines are presented in Appendix \ref{appendix:anno_demographic} and \ref{appendix:anno_guideline}.}

\paragraph{QA Pairs Creation}
We first instruct the annotators to create commonsense question-answer pairs based on the given category and question concept that we had newly created (\S\ref{sec:task_scope}). We also ask them to rely on their existing knowledge when making questions and avoid using internet search or LLMs as much as possible.
Each concept is annotated by 2 annotators, with each annotator covering all 5 categories to ensure consistency. Since we have 150 unique concepts per category and 6 annotators per language, each annotator is assigned to create QA pairs for 50 concepts per category.
This results in 300 QA pairs per category (150 concepts $\times$ 2 annotators) and a total of 1,500 QA pairs per language (300 QA pairs $\times$ 5 categories).

\paragraph{Answering Question}
After passing the quality assurance, the 1,500 QA pairs are then redistributed among all annotators for further review, where they are tasked with answering 1,250 commonsense questions each (excluding the set of data they have made in the first phase). Therefore, in this phase, each question is answered by five different annotators to capture the consistency and variance in the commonsense knowledge among annotators from various cultural backgrounds. Annotators are instructed not only to provide answers to the questions but also to comment on any ambiguities in the questions or options, or if they have any uncertainties when answering the questions.

\paragraph{Quality Control}
We conduct Quality Control (QC) through manual validation to ensure data accuracy and maintain its ``commonsense'' nature. QC annotators manually review data for errors and provide feedback, which is then corrected by the annotator concerned. Once corrected, QC annotators re-check the data to determine whether it can be considered complete or still requires revision. An evaluation meeting with all annotators is also held to convey a more comprehensive evaluation regarding all types of errors and other findings, ensuring annotators do not repeat similar errors and have a better understanding of the notion of commonsense. After the second phase (answering questions), we also excluded some questions if more than three annotators marked them as ambiguous.

\subsection{Automatic Data Generation}
\label{sec:llm_gen}
For generating the second type of LLM-generated data, rather than adapting questions from the English dataset (\S\ref{sec:llm_adapt}), we use the same set of categories and question concepts as the human-generated data (\S\ref{sec:human_gen}). We also utilize GPT-4 Turbo and instruct it to generate questions, options, and answers. To ensure the generated dataset aligns closely with the intended cultural context, we explicitly incorporated the categories and question concepts in the prompt. Additionally, we instruct the model to strictly include the question concepts in the generated questions.

We adopt a batching approach to streamline the data generation process, providing the model with a maximum of 5 distinct question concepts from the same categories in one API call. We chose 5 concepts because this number is optimal; larger batches would exceed the API's maximum length, while smaller batches would result in longer processing times.

To maintain dataset quality and uniqueness, we filter out duplicate entries and questions that do not explicitly contain the question concepts.

\section{Data Analysis}
\subsection{LLM-Generated Data}
\label{sec:llm_generated_data_analysis}
To evaluate the quality of our LLM-generated data, we manually reviewed all 158 samples from the \textsc{LLM\_Adapt} test set (see Table \ref{tab:data_stat}). Based on the evaluation, we then calculate the accuracy of the generated concepts, questions, and options. We also evaluate 300 randomly selected samples from the \textsc{LLM\_Gen} using the same procedure.


\begin{table}[t]
\centering
\small
\resizebox{0.95\columnwidth}{!}{%
\begin{tabular}{@{}lrll@{}}
\toprule
\multicolumn{1}{c}{\multirow{2}{*}{\textbf{Status}}} &
  \multicolumn{1}{c}{\multirow{2}{*}{\textbf{Num (\%)}}} &
  \multicolumn{2}{c}{\textbf{Concept Example}} \\ \cmidrule(l){3-4} 
\multicolumn{1}{c}{} &
  \multicolumn{1}{c}{} &
  \multicolumn{1}{c}{\textbf{Orig (\texttt{eng})}} &
  \multicolumn{1}{c}{\textbf{Modified (\texttt{ind})}} \\ \midrule
\multirow{2}{*}{\begin{tabular}[c]{@{}l@{}}Correct\\ (major)\end{tabular}} &
  \multirow{2}{*}{32 (20.25\%)} &
  beaver &
  \begin{tabular}[c]{@{}l@{}}komodo\end{tabular} \\ \cmidrule(l){3-4}
 &
   &
  snow &
  \begin{tabular}[c]{@{}l@{}}hujan abu vulkanik\\ (\textit{volcanic ashfall})\end{tabular} \\ \midrule
\multirow{2}{*}{\begin{tabular}[c]{@{}l@{}}Correct\\ (minor)\end{tabular}} &
  \multirow{2}{*}{119 (75.32\%)} &
  tower &
  menara (\textit{tower}) \\ \cmidrule(l){3-4}
 &
   &
  grape &
  anggur (\textit{grape}) \\ \midrule
\multirow{2}{*}{Wrong} &
  \multirow{2}{*}{7 (4.43\%)} &
  orchestra pit &
  \begin{tabular}[c]{@{}l@{}}\red{sumur} orkestra\\ (\textit{orchestra well})\end{tabular} \\ \cmidrule(l){3-4}
 &
   &
  skate &
  \red{ice skating} \\ \bottomrule
\end{tabular}%
}
\vspace{-4pt}
\caption{\texttt{eng} to \texttt{ind} concept adaptation result.}
\label{tab:concept_sample_ind}
\end{table}

\begin{table}[t]
\centering
\small
\resizebox{0.95\columnwidth}{!}{%
\begin{tabular}{@{}lrll@{}}
\toprule
\multicolumn{1}{c}{\multirow{2}{*}{\textbf{Status}}} &
  \multicolumn{1}{c}{\multirow{2}{*}{\textbf{Num (\%)}}} &
  \multicolumn{2}{c}{\textbf{Concept Example}} \\ \cmidrule(l){3-4} 
\multicolumn{1}{c}{} &
  \multicolumn{1}{c}{} &
  \multicolumn{1}{c}{\textbf{Orig (\texttt{ind})}} &
  \multicolumn{1}{c}{\textbf{Modified (\texttt{sun})}} \\ \midrule
\multirow{2}{*}{Correct} &
  \multirow{2}{*}{\begin{tabular}[c]{@{}r@{}}122\\ (77.22\%)\end{tabular}} &
  \begin{tabular}[c]{@{}l@{}}hujan abu vulkanik\\ (\textit{volcanic ashfall})\end{tabular} &
  \begin{tabular}[c]{@{}l@{}}hujan lebu vulkanik\\ (\textit{volcanic ashfall})\end{tabular} \\ \cmidrule(l){3-4} 
 &
   &
  menara (\textit{tower}) &
  munara (\textit{tower}) \\ \midrule
\multirow{2}{*}{Wrong} &
  \multirow{2}{*}{\begin{tabular}[c]{@{}r@{}}36\\ (22.78\%)\end{tabular}} &
  \begin{tabular}[c]{@{}l@{}}cicak\\ (\textit{house gecko})\end{tabular} &
  \begin{tabular}[c]{@{}l@{}}\red{kadal imah}\\ (\textit{house lizard})\end{tabular} \\ \cmidrule(l){3-4} 
 &
   &
  \begin{tabular}[c]{@{}l@{}}klinik gigi\\ (\textit{dental clinic})\end{tabular} &
  \begin{tabular}[c]{@{}l@{}}klinik \red{dental}\\ (\textit{dental clinic})\end{tabular} \\ \bottomrule
\end{tabular}%
}
\vspace{-4pt}
\caption{\texttt{ind} to \texttt{sun} concept adaptation result.}
\label{tab:concept_sample_sun}
\end{table}
\subsubsection{Concept Analysis}
\paragraph{Concept Quality}
To evaluate the quality of the adapted concepts, we regard the concept as correct if it is a real, existing concept (not a hallucination) and relevant to the Indonesian or Sundanese context. As shown in Table \ref{tab:concept_sample_ind}, 95.57\% of English to Indonesian adaptations are correct, including 16 out of 19 concepts needing major adaptation, such as `snow' to `\textit{hujan abu vulkanik}' (volcanic ashfall). This result indicates that the LLM can adequately adapt some English concepts to Indonesian, although most involve direct translations without major alterations. However, for Indonesian to Sundanese (Table \ref{tab:concept_sample_sun}), the correct adaptation drops to 77.22\%, reflecting weak machine translation (MT) performance in Sundanese.

\begin{figure}[t!]
    \begin{center}
        \centerline{\includegraphics[width=0.95\columnwidth]{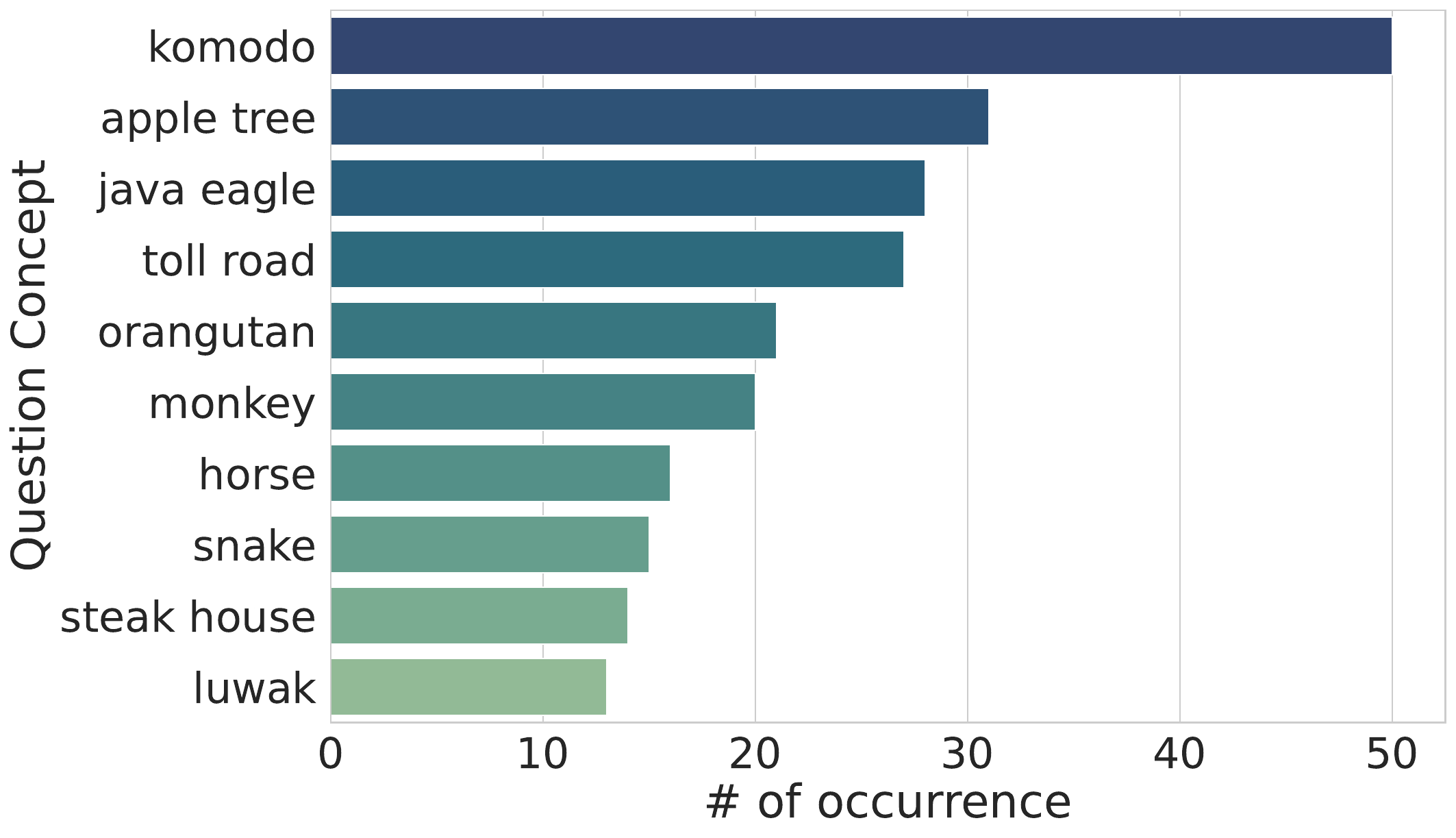}}
        \vspace{-8pt}
        \caption{Top-10 adapted question concepts taken from train, validation, and test set of \textsc{LLM\_Adapt} data.}
        \vspace{-8pt}
        \label{fig:concept_top_10}
    \end{center}
\end{figure}

\paragraph{Concept Variation}
Despite the high accuracy of concept adaptation, as shown in Figure \ref{fig:concept_top_10}, the adapted concepts are skewed towards \textit{`komodo,'} indicating a bias towards a specific entity within a category. This could be due to the model being trained on data with insufficient knowledge or the absence of a direct equivalent of the English concept in Indonesian, leading it to default to one standard concept. This finding highlights the limitations of concept adaptation from existing English data. To improve dataset diversity and coverage, manual concept creation in the target language is needed.

\subsubsection{Question Analysis}
\paragraph{Question Quality}
To evaluate the quality of the generated questions, we apply a \textit{strict} criterion: any errors, even minor, are marked as incorrect. As shown in Table \ref{tab:question_acc_ind_sun}, the Indonesian datasets have a high percentage of error-free questions, ranging between 68--75\%. However, for Sundanese, the accuracy decreases significantly. In particular, the weak performance of the MT system in Sundanese is evident from its very low accuracy in \textsc{LLM\_Adapt}, suggesting synthetic data adaptation/translation from English is not an optimal method for low-resource languages. In contrast, the \textsc{LLM\_Gen} dataset, which involves generating synthetic data directly in the target language, shows a significantly higher number of correct questions, particularly for Sundanese. This indicates that direct generation in the target language is a more promising approach than adaptation or translation.

\paragraph{Common Mistakes}
We manually reviewed the questions to identify common errors, detailed in Table \ref{tab:question_errors_ind_sun}. We observe that most of the errors in \textsc{LLM\_Adapt} come from translation errors. For Indonesian \textsc{LLM\_Gen}, despite lower question generation accuracy, most of the errors are actually minor typos. In Sundanese \textsc{LLM\_Gen}, however, the predominant issues relate to sentence fluency, indicating ongoing challenges in automatically generating smooth and fluent sentences in Sundanese. We also found that the model occasionally outputs wrong languages, particularly using Indonesian phrases or words instead of Sundanese. While Indonesian and Sundanese share many linguistic features, they differ significantly in morphological features (\S\ref{sec:languages_in_indonesia}). This difference, along with the smaller amount of Sundanese data used for training LLMs, results in the models being less exposed to Sundanese words, suffixes, or prefixes compared to Indonesian. This leads to the occasional appearance of Indonesian words in generated Sundanese data, even though the sentences may be grammatically correct.

\subsubsection{Options and Answer Analysis}
\begin{table}[t]
\centering
\small
\begin{tabular}{@{}lcc@{}}
\toprule
\multicolumn{1}{c}{\multirow{2}{*}{\textbf{Dataset}}} & \multicolumn{2}{c}{\textbf{\% of correct questions}}                                    \\ \cmidrule(l){2-3} 
\multicolumn{1}{c}{}                                 & \multicolumn{1}{c}{\textbf{\texttt{ind}}} & \multicolumn{1}{c}{\textbf{\texttt{sun}}} \\ \midrule
\textsc{LLM\_Adapt}                & \textbf{75.32\%}                            & 15.19\%                     \\
\textsc{LLM\_Gen}                 & 68.67\%                            & \textbf{51.00\%}                     \\ \bottomrule
\end{tabular}
\vspace{-4pt}
\caption{Question generation acc. of LLM-generated datasets, measured by the \% of error-free questions.}
\label{tab:question_acc_ind_sun}
\end{table}

\begin{table}[t]
\centering
\small
\resizebox{\columnwidth}{!}{%
\begin{tabular}{@{}lrrrr@{}}
\toprule
\multicolumn{1}{c}{\multirow{3}{*}{\textbf{Error Type}}} & \multicolumn{4}{c}{\textbf{\% of questions}} \\ \cmidrule(l){2-5} 
\multicolumn{1}{c}{}& \multicolumn{2}{c}{\textbf{\textsc{LLM\_Adapt}}} & \multicolumn{2}{c}{\textbf{\textsc{LLM\_Gen}}}\\ \cmidrule(l){2-3} \cmidrule(l){4-5}
\multicolumn{1}{c}{}& \multicolumn{1}{c}{\textbf{\texttt{ind}}} & \multicolumn{1}{c}{\textbf{\texttt{sun}}} & \multicolumn{1}{c}{\textbf{\texttt{ind}}} & \multicolumn{1}{c}{\textbf{\texttt{sun}}} \\ \midrule
\textit{No error}& \textit{75.32\%}  & \textit{15.19\%}  &  \textit{68.67\%}  & \textit{51.00\%} \\ \cdashlinelr{1-5}
Translation & \textbf{8.23\%}  & \textbf{41.14\%}  &  0.00\%  & 0.00\% \\
Wrong language & 0.00\%  & 0.00\%  &  0.00\%  & 6.67\% \\
Sent. structure  & 3.16\%  & 10.13\%  &  0.00\%  &  0.00\%  \\
Sent. fluency & 6.96\%  & 23.42\%  & 11.33\%  & \textbf{18.00\%} \\
Sent. context &  1.90\%  &  1.90\%  &  3.00\%  & 8.00\% \\
Subjectivity  &  0.63\%  &  0.63\%  &  0.00\%  &  0.00\%  \\
Typo/mechanics& 3.80\%  &  7.59\%  & \textbf{17.00\%}  & 16.33\% \\ \bottomrule
\end{tabular}
}
\vspace{-4pt}
\caption{Distribution of question generation error types of LLM-generated datasets.}
\label{tab:question_errors_ind_sun}
\end{table}
\begin{table}[t]
\centering
\small
\begin{tabular}{@{}lcc@{}}
\toprule
\multicolumn{1}{c}{\multirow{2}{*}{\textbf{Dataset}}} & \multicolumn{2}{c}{\textbf{\% of correct options}}                                    \\ \cmidrule(l){2-3} 
\multicolumn{1}{c}{}                                 & \multicolumn{1}{c}{\textbf{\texttt{ind}}} & \multicolumn{1}{c}{\textbf{\texttt{sun}}} \\ \midrule
\textsc{LLM\_Adapt}                & 62.66\%                            & 38.61\%                     \\
\textsc{LLM\_Gen}                 & \textbf{93.00\%}                            & \textbf{58.67\%}                     \\ \bottomrule
\end{tabular}
\vspace{-4pt}
\caption{Options generation acc. of LLM-generated datasets, measured by the \% of error-free options.}
\label{tab:choices_acc_ind_sun}
\end{table}

\begin{table}[t]
\centering
\small
\resizebox{\columnwidth}{!}{%
\begin{tabular}{@{}lrrrr@{}}
\toprule
\multicolumn{1}{c}{\multirow{3}{*}{\textbf{Error Type}}} & \multicolumn{4}{c}{\textbf{\% of options}} \\ \cmidrule(l){2-5} 
\multicolumn{1}{c}{}& \multicolumn{2}{c}{\textbf{\textsc{LLM\_Adapt}}} & \multicolumn{2}{c}{\textbf{\textsc{LLM\_Gen}}}\\ \cmidrule(l){2-3} \cmidrule(l){4-5}
\multicolumn{1}{c}{}& \multicolumn{1}{c}{\textbf{\texttt{ind}}} & \multicolumn{1}{c}{\textbf{\texttt{sun}}} & \multicolumn{1}{c}{\textbf{\texttt{ind}}} & \multicolumn{1}{c}{\textbf{\texttt{sun}}} \\ \midrule
\textit{No error}& \textit{62.66\%}  & \textit{38.61\%}  &  \textit{93.00\%}  & \textit{58.67\%} \\ \cdashlinelr{1-5}
Translation & 3.80\%  & \textbf{47.47\%}&  0.00\%  & 0.00\%   \\
Wrong language & 0.00\%  & 0.00\% &  0.00\%  & \textbf{31.67\%}   \\
Sent. fluency&  0.63\%  &  0.63\%&  0.67\%  &  1.33\%   \\
Sent. context&  0.63\%  &  0.63\%&  0.00\%  &  0.00\%   \\
Invalid options&  1.27\%  &  1.27\%& \textbf{5.67\%}  & 8.33\%   \\ 
Typo/mechanics  & \textbf{31.01\%}  & 11.39\%&  0.67\%  &  0.00\%   \\
\bottomrule
\end{tabular}
}
\vspace{-4pt}
\caption{Distribution of options generation error types of LLM-generated datasets.}
\label{tab:choices_errors_ind_sun}
\end{table}

\paragraph{Options Quality}
We also assess the options quality using a similar method to the question evaluation. From Table \ref{tab:choices_acc_ind_sun}, we find \textsc{LLM\_Gen} generates higher quality options compared to \textsc{LLM\_Adapt}. However, there is still a significant gap in performance between Indonesian and Sundanese, once again highlighting the performance discrepancy between medium- and lower-resource languages.

\paragraph{Common Mistakes}
As detailed in Table \ref{tab:choices_errors_ind_sun}, for \textsc{LLM\_Adapt}, the most common issues are minor typos and mechanical errors, particularly with capitalization (e.g., "indonesia" instead of "Indonesia"). For Sundanese, the errors are mainly due to major translation errors. As for \textsc{LLM\_Gen}, the most common errors involve the presence of invalid options, particularly when no correct answers are among the options. For Sundanese, alongside invalid options, the model also produces some options in the wrong language. Still, it is important to note that the proportion of invalid options is very low for both languages, indicating that the model generates a substantial number of valid options.

\subsection{LLM vs. Human: Lexical Diversity}
To compare the lexical diversity between LLM and human-generated data, we analyze the proportion of shared tokens between the \textsc{LLM\_Gen} and \textsc{Human\_Gen}, calculated by dividing the number of unique shared tokens by the total number of unique tokens. We find that the unigram overlap percentage is 39.75\% for \textsc{Human\_Gen} and 65.48\% for \textsc{LLM\_Gen}. A similar trend is observed for the bigram overlap percentage, with 12.41\% for \textsc{Human\_Gen} and 15.98\% for \textsc{LLM\_Gen}. This shows that many tokens present in \textsc{Human\_Gen} also exist in \textsc{LLM\_Gen}, but the reverse is not equally true. We also find that \textsc{Human\_Gen} dataset has 8,596 unique tokens, higher than \textsc{LLM\_Gen} with 6,677 unique tokens.

Upon sample-level analysis, we find that given the same set of categories and question concepts, humans generate more token variations that are not produced by LLMs, such as some unique terms like \textit{kalis} or \textit{cimol.}\footnote{In the context of culinary, \textit{`kalis'} means a state of dough that is well-kneaded and ready to be processed further. \textit{`Cimol'} is an Indonesian street food made from tapioca flour.} Although LLM can still generate some valid data, it tends to return more popular concepts or general questions. For instance, given \textit{`kerupuk'} (crackers) concept, human annotators can create questions tied to their cultural background, such as asking about \textit{`kerupuk rambak'} (rambak crackers). In contrast, LLM tends to give more general (but still relevant) questions, like \textit{"What is the common primary ingredient of crackers?"} More examples are shown in Table \ref{tab:llm_vs_human_gen_examples} in the Appendix.

\section{Benchmark Result}
\subsection{Experiment Setup}
We conduct a zero-shot evaluation of various LLMs to assess their performance on our datasets.

\vspace{-4px}
\paragraph{English-centric LLMs} We include LLaMA-2 7B and 13B \cite{touvron2023llama}, a widely used open LLM, and MistralOrca-7B \cite{lian2023mistralorca}.

\vspace{-4px}
\paragraph{Multilingual LLMs} We include PolyLM-13B \cite{wei2023polylm}, which trained on a multilingual dataset (mostly English and Chinese); BLOOMZ-7B \cite{muennighoff2022crosslingual}, which fine-tuned on xP3; SeaLLM-7B \cite{nguyen2023seallms}, which covers Southeast Asian languages; and Aya \cite{ustun2024aya}, recent open LLM trained on 101 languages, including Indonesian and Sundanese.

\vspace{-4px}
\paragraph{Monolingual LLMs} We include Merak-v4 \cite{Merak}, an Indonesian LLM based on MistralOrca and MalayMistral-7B \cite{zolkepli2024large}, a Malaysian LLM extended from Mistral.

\vspace{-4px}
\paragraph{Proprietary LLMs} We include GPT-3.5 Turbo, GPT-4, and GPT-4 Turbo \cite{openai2023gpt4}.

\vspace{4px}
We use three prompt variations for all models.\footnote{Please refer to Appendix \ref{appendix:benchmark_prompt} for the prompts details.} For a fair comparison between open and proprietary LLMs, we extract the answer key from the text generation result instead of the next token probability, using a rule-based and regex. Evaluations are performed on RTX A6000 48GB. We use accuracy as the evaluation metric.

\subsection{Overall Performance}
\begin{figure}[t!]
    \begin{center}
    \centerline{\includegraphics[width=\columnwidth]{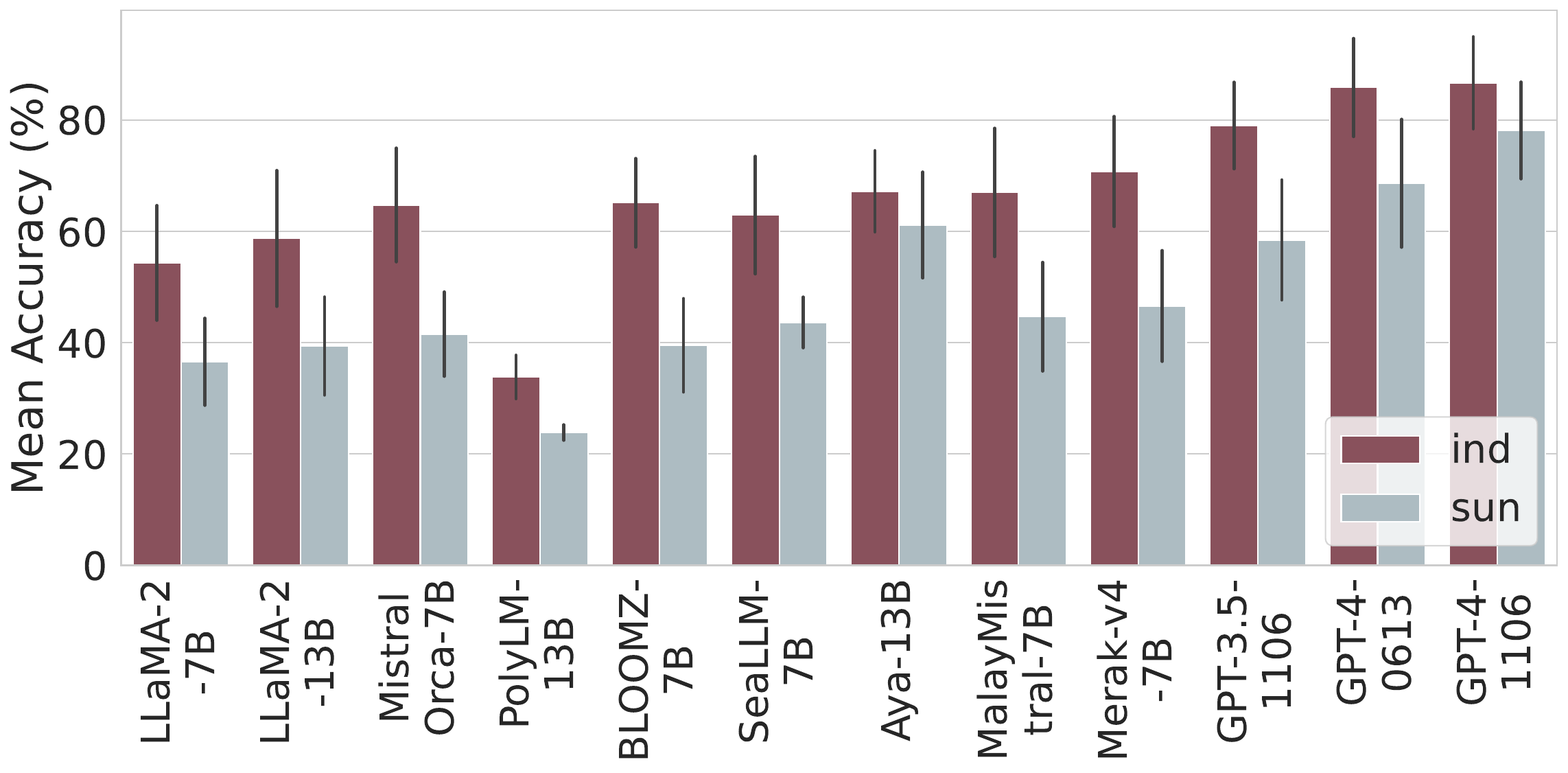}}
        \vspace{-6pt}
        \caption{LLMs' performance on our combined test set.}
        \label{fig:overall_benchmark}
    \end{center}
\end{figure}

We first benchmark all selected LLMs on our combined datasets to measure the overall performance. As shown in Figure \ref{fig:overall_benchmark}, GPT models outperform other LLMs, with an average $\sim$80\% accuracy. Among open models, Merak-v4 scores highest in Indonesian but does not surpass GPT-3.5. Interestingly, the score difference between Merak-v4 and MalayMistral is small, possibly due to the high lexical similarities between Indonesian and Standard Malay. This may also be due to some instruction data used to train MalayMistral is generated using GPT-4, which tends to produce texts in Indonesian rather than Standard Malay.

We also observe a substantial gap between Indonesian and Sundanese (10--20\% accuracy drop), suggesting that current LLMs struggle with Sundanese questions, even in multiple-choice settings. This gap, particularly in Merak-v4, highlights the limitations of training LLMs solely on Indonesian texts, which does not ensure transferable performance across other local languages due to morphological differences. Including Sundanese texts in the training data, as shown by the improved performance in Aya, effectively narrows the performance gap between Indonesian and Sundanese.

\subsection{LLM vs. Human-Generated Data}
\begin{figure}[t!]
    \begin{center}
        \vspace{-12px}
        \centerline{\includegraphics[width=\columnwidth]{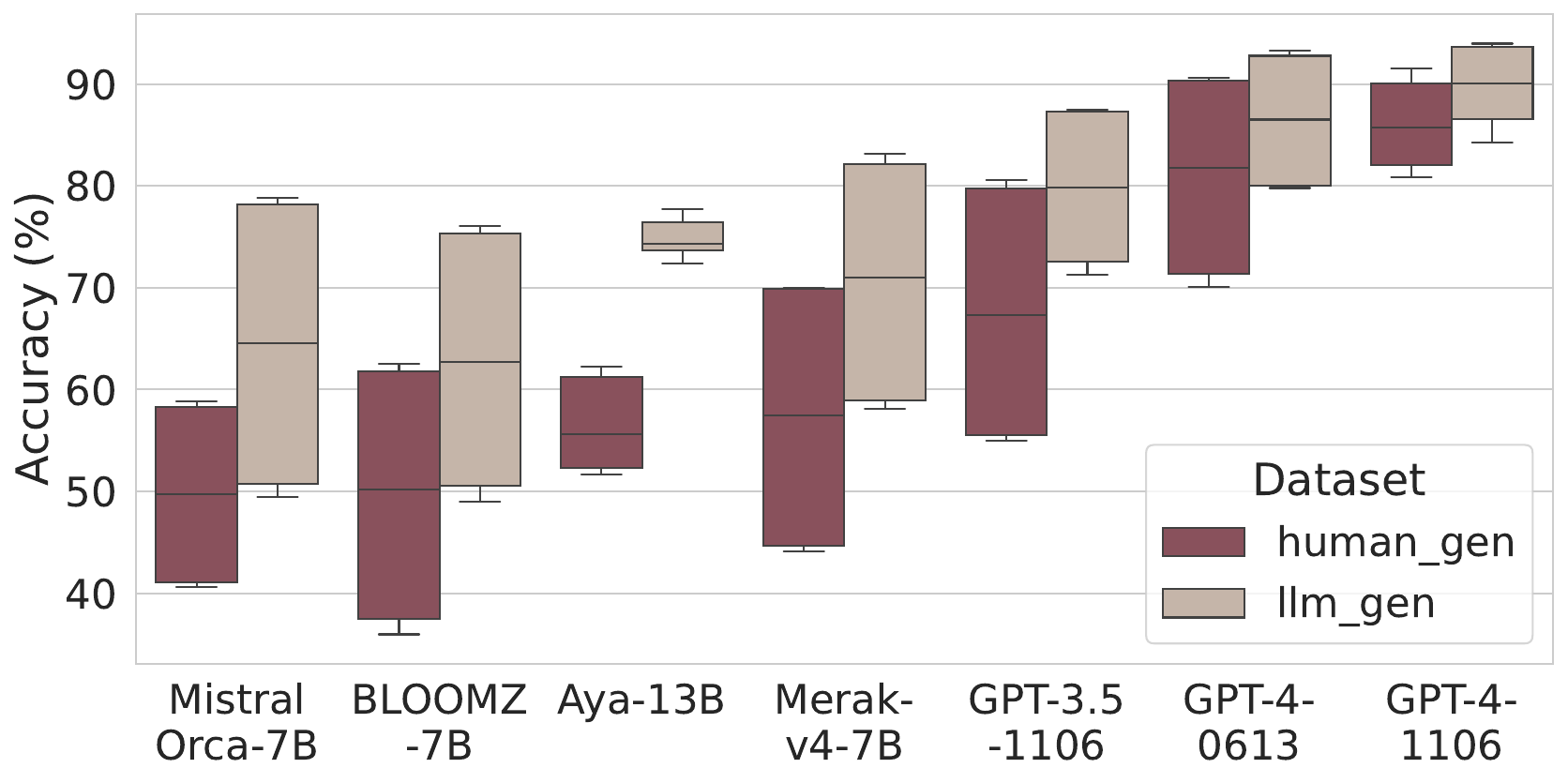}}
        \vspace{-6px}
        \caption{LLMs' performance on \textsc{LLM\_Gen} vs. \textsc{Human\_Gen} in Indonesian and Sundanese. We combined data points from both languages for visualization, with lower quartiles typically representing Sundanese data.}
        \vspace{-10px}
        \label{fig:open_closed_gap}
    \end{center}
\end{figure}

To assess LLMs' ability to answer LLM and human-generated data, we compare their performances on our \textsc{LLM\_Gen} and \textsc{Human\_Gen} datasets. Figure \ref{fig:open_closed_gap} shows that LLMs generally perform higher on \textsc{LLM\_Gen}, especially on MistalOrca, which is trained on English-centric data. This suggests the model can handle \textsc{LLM\_Gen} questions despite not being specifically trained on Indonesian or Sundanese data. Still, it is important to note that MistalOrca's average score on \textsc{LLM\_Gen} is still lower than other LLMs that include Indonesian texts in their training.

Interestingly, the performance gap tends to narrow as models improve, particularly for GPT models. For Aya, despite relatively strong performance compared to other open-source LLMs, shows significantly dropped performance in \textsc{Human\_Gen} data. This decline is likely due to Aya being a `translation-heavy' model \cite{ustun2024aya}, with a large proportion of its training data derived from translations rather than human annotations.

\begin{figure*}[t!]
    \begin{center}
        \centerline{\includegraphics[width=\textwidth]{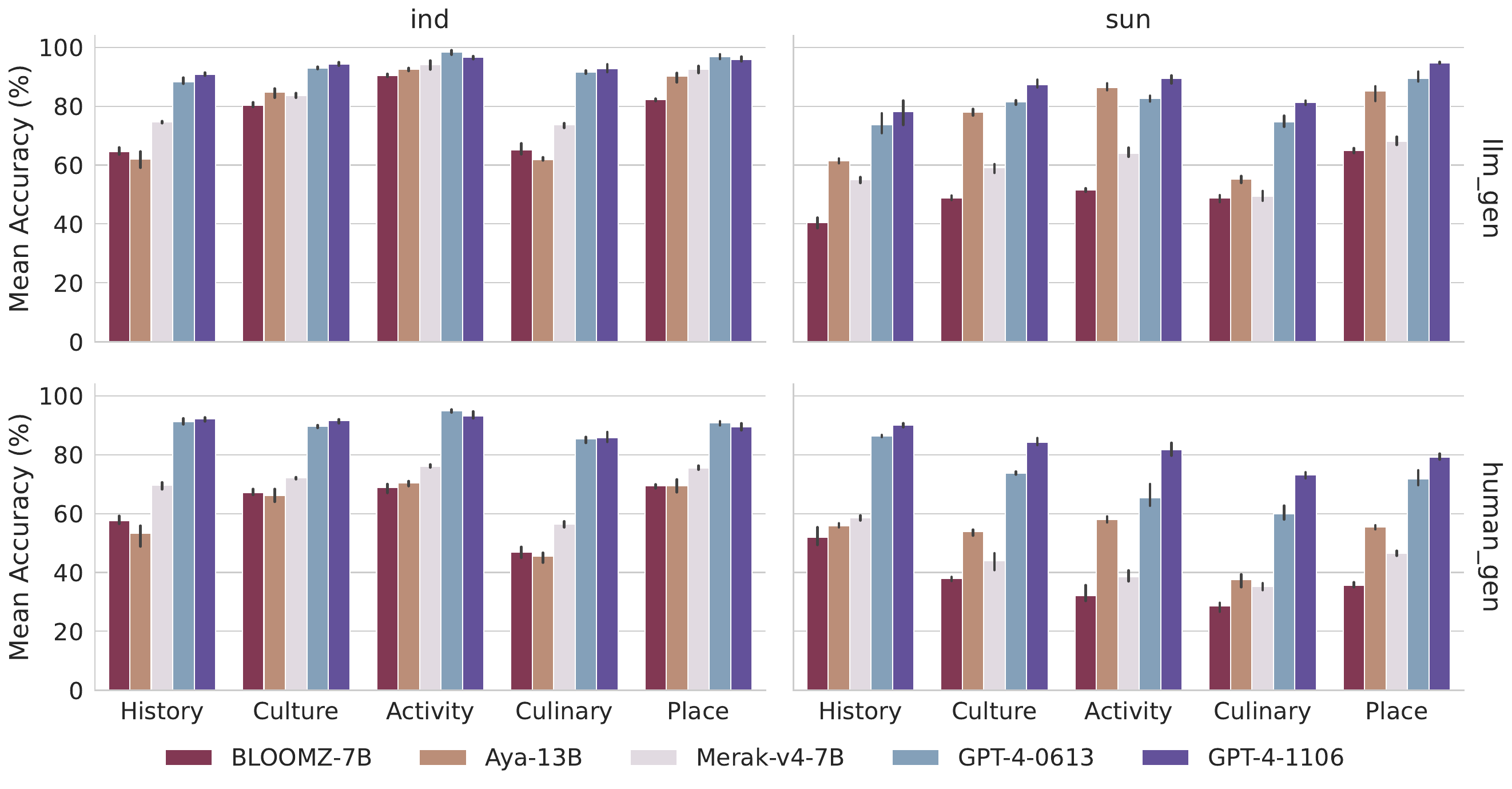}}
        \vspace{-8px}
        \caption{LLMs performance by question category in \textsc{LLM\_Gen} and \textsc{Human\_Gen} for Indonesian and Sundanese.}
        \label{fig:category_benchmark}
    \end{center}
\end{figure*}

\subsection{Performance by Question Category}
Figure \ref{fig:category_benchmark} shows the LLMs' performance across different question categories. It shows that LLMs perform better in the \textit{activity} and \textit{place} questions, which benefit from a lot of information readily available on the internet. However, they struggle with \textit{culinary} questions, where specialized or cultural knowledge is often required. Interestingly, a significant improvement is seen from GPT-4 (\texttt{0613} ver.) to GPT-4 Turbo (\texttt{1106} ver.), particularly in Sundanese \textsc{Human\_Gen}. This suggests that the model is ``acquiring'' more knowledge, possibly due to the interactions on the ChatGPT web. However, some categories still score below 80\%, showing there is still room for improvement.

\section{Discussion}
\subsection{Multiple-Choice vs. `Free' Generation}
\label{sec:multiple_choice_vs_free}
Since our dataset is in a multiple-choice format, LLMs might look better than they actually are because they can just pick one answer from the given options. To see LLMs' actual capability, we test GPT-4 Turbo on 100 randomly sampled questions from Indonesian \textsc{Human\_Gen} in open-ended settings, i.e., asking the question directly without showing answer options or providing any extra instructions. Our manual evaluation shows an accuracy of 75\%, a huge drop from the $\sim$90\% accuracy in multiple-choice settings. This highlights the limitation of the model in open-ended settings.

Among the wrong answers, most cases come from overly general answers not specific to Indonesia, especially for questions in \textit{culinary}, \textit{culture}, and \textit{activity} categories.
For instance, given unique \textit{activity} questions such as: \textit{"What is the mandatory song to be sung during the ‘mengheningkan cipta’ (moment of silence) in the flag ceremony?"} the model gives a wrong answer in open-ended generation settings even if it could identify the correct answer in the multiple-choice setup. More failure examples can be seen in Table \ref{tab:open_gen_examples} in the Appendix.

\subsection{LLM-generated Data Quality}
Due to the high performance of LLMs in answering LLM-generated data, one might argue that such data is less valuable. Ideally, creating a larger human-generated dataset is the most straightforward method to ensure data quality. However, this approach may not always be feasible in the context of underrepresented languages, particularly given the limited resources. Building a dataset manually can be costly and challenging (we spent more than \$3,000 in total to build \textsc{Human\_Gen}), and leveraging LLMs could be a practical and cheaper alternative, especially for scaling up the data.

To address the limitation of LLM-generated data, we can apply additional processes to reduce the noise. For instance, a collaboration between humans and models can be conducted to fix and revise the synthetic questions \cite{liu-etal-2022-wanli,putri-oh-2022-idk}. This noise can also be reduced automatically, especially in Indonesian \textsc{LLM\_Gen} questions, where errors typically involve minor typos.

It is also important to note that the quality of LLM-generated data can vary. Our experiments show \textsc{LLM\_Gen} has significantly fewer errors than \textsc{LLM\_Adapt}, particularly in Sundanese. We also compared the performance between the cleaned vs. raw version of the data (Appendix \ref{appendix:data_cleaning}) and found that cleaning the data had a relatively minor impact on performance in \textsc{LLM\_Gen}. Still, it remains crucial not to depend solely on LLMs, as they still have limitations, such as producing convincing yet incorrect outputs \cite{pan-etal-2023-risk}.

\section{Conclusion}
In this study, we introduced new Indonesian and Sundanese CommonsenseQA datasets built using various methods: automatic dataset adaptation, direct generation with LLMs, and manual dataset creation by human annotators from diverse regions. Our thorough analysis reveals that adapting existing English data is less effective for Sundanese. In contrast, direct generation in the target languages by GPT-4 Turbo produces relevant data for both languages, although the cultural depth has not yet matched the human-generated ones. Nevertheless, given the limited resources, combining LLMs and humans for dataset creation can be a practical solution, allowing for more efficient dataset creation by reducing the over-reliance on a single data source.

Our goal is for LLMs to serve as beneficial tools for diverse communities, not solely those from higher-resource languages, cultures, or regions. This is especially important since LLMs are continuously being used to generate data. We hope our work represents a crucial step toward achieving this broader objective.

\section*{Limitations}
\paragraph{Language and Region Coverage}
In terms of language coverage, we were only able to cover Indonesian and Sundanese due to the available resources and the authors' familiarity with these languages. Additionally, the annotators we recruited were mostly from Java island, with one annotator from Bali island. Despite our effort to include a range of question concepts from different regions, including those beyond Java and Bali islands, it is possible that some bias may exist, especially in the Indonesian dataset. This is because the questions were generated primarily by annotators from Java and Bali, and their perspectives and cultural backgrounds may have influenced the content. Nonetheless, we have taken measures to eliminate potentially harmful or stereotypical questions.

\paragraph{Extension to Other Local Languages}
Besides Indonesian, our study focuses on one Indonesian local language, Sundanese. As previously discussed, Indonesia has many local languages; however, we cannot cover all of them due to resource constraints. We aim for our findings in Sundanese to act as a starting point for other languages. We anticipate that LLMs might perform worse than Sundanese for extremely low-resource languages, like Buginese or Toba Batak. Javanese, on the other hand, is expected to have comparable performance to Sundanese \cite{winata-etal-2023-nusax,bang-etal-2023-multitask}. Generating synthetic data could be particularly beneficial for extremely low-resource languages, given the difficulty of finding native speakers. Yet, our case study in Sundanese indicates that LLMs might struggle even more with generating fluent sentences and/or culturally unique concepts in such languages. Still, we believe that the combination or even collaboration between LLM and humans may serve as a starting point.

\paragraph{Multiple-Choice Format}
We use a multiple choice question format, following English CommonsenseQA data format \cite{talmor-etal-2019-commonsenseqa} to facilitate a more straightforward and robust evaluation process. Although open-ended generation may offer a more challenging benchmark for LLMs, as also discussed in Section \ref{sec:multiple_choice_vs_free}, evaluating LLMs in such settings poses its own set of challenges, especially in low-resource languages where `LLM-as-a-judge' approach may not be as effective as in English. Nevertheless, our dataset can serve as a starting point for this line of research direction. 

\paragraph{Usage of GPT-4 for \textsc{LLM\_Gen}}
\textsc{LLM\_Gen} can initially be seen as a dataset to ``only'' close the gap between local LLMs and GPT-4 since we solely use GPT-4 as the data generator. At the time of the experiment (October--December 2023), using GPT-4 was our only sufficient option. Moving forward, with the rise of better LLMs that support Indonesian, such as Command-R+\footnote{\url{https://docs.cohere.com/docs/command-r-plus}} or Claude 3.5 Sonnet\footnote{\url{https://www.anthropic.com/news/claude-3-5-sonnet}}, we can enrich the existing data by including the generation of multiple ``strong'' LLMs. This will make \textsc{LLM\_Gen} a more comprehensive testing tool. Additionally, we would like to emphasize that even if \textsc{LLM\_Gen} is seen ``only'' as a tool for measuring the gap between local models and GPT-4, we believe its value should not be underestimated. Considering the lack of strong-performing open-source LLMs in low-resource languages like Sundanese, achieving comparable performance to GPT-4 in these languages would be highly valuable for the community.

\paragraph{Dataset Usage}
\textsc{Human\_Gen} and \textsc{LLM\_Gen} datasets should be used exclusively as test data. When utilizing the training data portion in the \textsc{LLM\_Adapt} dataset, special caution is required, particularly for the Sundanese language, where numerous errors have been identified. Training models on this data have a risk of error propagation of incorrect information. Additionally, the test data portions in both \textsc{LLM\_Adapt} and \textsc{LLM\_Gen} should be used carefully, as they contain ``silver'' labels and may include some inaccuracies. We strongly advise using the cleaned version of the data available on our GitHub repository and accompanying the evaluation results of the LLM-generated data with those from the \textsc{Human\_Gen} dataset. This is because high performance of the evaluated model, especially on the LLM-generated data, does not necessarily indicate the robustness of the model. One should be careful when making claims.

\section*{Ethical Consideration}
All human-generated datasets have been manually validated to ensure that harmful or offensive questions are not present in the dataset. We also excluded potentially harmful questions in the LLM-generated datasets through automatic filtering. Our work has been reviewed by KAIST Institutional Review Board (IRB). All recruited annotators were paid above the minimum wage based on the standard in Republic of Korea. Our datasets are publicly available under the Creative Commons Non-Commercial (CC BY-NC 4.0) license.

\section*{Acknowledgements}
This work was supported by Institute of Information \& communications Technology Planning \& Evaluation (IITP) grant funded by the Korea government(MSIT) (No. RS-2024-00509258, AI Guardians: Development of Robust, Controllable, and Unbiased Trustworthy AI Technology).

\bibliography{anthology,custom}

\appendix

\lstset{
  basicstyle=\small\tt,
  basewidth=4.5pt,
  breakindent=0em,
  breaklines=true,
}

\newenvironment{prompt}
  {\quoting[leftmargin=0pt, rightmargin=0pt]%
   \small 
   \ttfamily 
   }
  {\endquoting}


\section{LLM-Generated Data}
\subsection{Prompts Details}
\label{appendix:data_gen_prompts}
\subsubsection{Concept Relevancy Classifier}
\label{appendix:relevancy_classifier}
To classify whether a concept is relevant to the Indonesian and Sundanese context, we prompt GPT-3.5 Turbo\footnote{We used \texttt{gpt-3.5-turbo} API checkpoint, accessed in September 2023.} using an ensemble of five prompts. We condition the prompts based on whether the question concept is a verb or not, determined using the ConceptNet API. We set the location context as \textit{`Indonesia'} for Indonesian data and \textit{`West Java'} for Sundanese data. The examples of concept relevancy classifier results can be seen in Table \ref{tab:concept_classification_example}.

Below are the prompts if the question concept is a verb (e.g., \textit{skiing}, \textit{studying}):

\begin{flushleft}
    \textbf{Prompt Variation 1}
\end{flushleft}
\begin{prompt}
\begin{lstlisting}
Text: {QUESTION}
Concept: {QUESTION CONCEPT}

Can one '{QUESTION CONCEPT}' in {Indonesia/West Java}? Answer with 'yes' or 'no'.
\end{lstlisting}
\end{prompt}

\begin{flushleft}
    \textbf{Prompt Variation 2}
\end{flushleft}
\begin{prompt}
\begin{lstlisting}
Text: {QUESTION}
Concept: {QUESTION CONCEPT}

Do people in {Indonesia/West Java} familiar with '{QUESTION CONCEPT}' concept? Answer with 'yes' or 'no'.
\end{lstlisting}
\end{prompt}

\begin{flushleft}
    \textbf{Prompt Variation 3}
\end{flushleft}
\begin{prompt}
\begin{lstlisting}
Text: {QUESTION}
Concept: {QUESTION CONCEPT}

Is '{QUESTION CONCEPT}' concept exist in {Indonesia/West Java}? Answer with 'yes' or 'no'.
\end{lstlisting}
\end{prompt}

\begin{flushleft}
    \textbf{Prompt Variation 4}
\end{flushleft}
\begin{prompt}
\begin{lstlisting}
Text: {QUESTION}
Concept: {QUESTION CONCEPT}

In {Indonesia/West Java}, can people '{QUESTION CONCEPT}'? Answer with 'yes' or 'no'.
\end{lstlisting}
\end{prompt}

\begin{flushleft}
    \textbf{Prompt Variation 5}
\end{flushleft}
\begin{prompt}
\begin{lstlisting}
Text: {QUESTION}
Concept: {QUESTION CONCEPT}

Can '{QUESTION CONCEPT}' be done in {Indonesia/West Java}? Answer with 'yes' or 'no'.
\end{lstlisting}
\end{prompt}

Below are the prompts if the question concept is \textit{not} a verb (e.g., \textit{snow}, \textit{bald eagle}):

\begin{flushleft}
    \textbf{Prompt Variation 1}
\end{flushleft}
\begin{prompt}
\begin{lstlisting}
Text: {QUESTION}
Concept: {QUESTION CONCEPT}

Can one find '{QUESTION CONCEPT}' in {Indonesia/West Java}? Answer with 'yes' or 'no'.
\end{lstlisting}
\end{prompt}

\begin{flushleft}
    \textbf{Prompt Variation 2}
\end{flushleft}
\begin{prompt}
\begin{lstlisting}
Text: {QUESTION}
Concept: {QUESTION CONCEPT}

Do people in {Indonesia/West Java} familiar with '{QUESTION CONCEPT}' concept? Answer with 'yes' or 'no'.
\end{lstlisting}
\end{prompt}

\begin{flushleft}
    \textbf{Prompt Variation 3}
\end{flushleft}
\begin{prompt}
\begin{lstlisting}
Text: {QUESTION}
Concept: {QUESTION CONCEPT}

Is '{QUESTION CONCEPT}' concept exist in {Indonesia/West Java}? Answer with 'yes' or 'no'.
\end{lstlisting}
\end{prompt}

\begin{flushleft}
    \textbf{Prompt Variation 4}
\end{flushleft}
\begin{prompt}
\begin{lstlisting}
Text: {QUESTION}
Concept: {QUESTION CONCEPT}

In {Indonesia/West Java}, can people find '{QUESTION CONCEPT}'? Answer with 'yes' or 'no'.
\end{lstlisting}
\end{prompt}

\begin{flushleft}
    \textbf{Prompt Variation 5}
\end{flushleft}
\begin{prompt}
\begin{lstlisting}
Text: {QUESTION}
Concept: {QUESTION CONCEPT}

Can '{QUESTION CONCEPT}' be found in {Indonesia/West Java}? Answer with 'yes' or 'no'.
\end{lstlisting}
\end{prompt}

\subsubsection{Automatic Data Adaptation}
\label{appendix:llm_adapt_prompt}
There are two prompts used to adapt the CommonsenseQA into \textsc{LLM\_Adapt}: Adapt All and Adapt Name. Below are the details of both prompts.

\begin{flushleft}
    \textbf{Adapt All Prompt}
\end{flushleft}

\begin{prompt}
\begin{lstlisting}
Change the given data to make it relevant to Indonesia in any ways. Make all elements relevant to each other, and the concept always appear explicitly in the question. Return in Indonesian language with JSON format where question is string, concept is string, options is dictionary where label is the keys and option text is the values, and question_answer is string contain one label from the options.

Data:
###
Question: {QUESTION}
Concept: {QUESTION CONCEPT}
Options:
{CHOICES}
Question Answer: {ANSWER}
###
\end{lstlisting}
\end{prompt}

\begin{flushleft}
    \textbf{Adapt Name Prompt}
\end{flushleft}
\begin{prompt}
\begin{lstlisting}
Change all names in the given question to Indonesian names. Change only the names, keep all other phrases in the question the same and keep it all in Indonesian.

Question: {QUESTION}
Changed Question:
\end{lstlisting}
\end{prompt}

\subsubsection{Automatic Data Generation}
\label{appendix:llm_gen_prompt}
For \textsc{LLM\_Gen}, we directly generate the data given a set of question concepts. Below are the prompt used for data generation.

\begin{flushleft}
    \textbf{Data Generation Prompt}
\end{flushleft}
\begin{prompt}
\begin{lstlisting}
Given a list of {LANGUAGE} concepts [QUESTION CONCEPTS}], create one {LANGUAGE} commonsense QA data with topic "{CATEGORY}" for each concept, that consists of three components: "question", "choices", and "answer_creator". The "question" must contains the concept explicitly. The "choices" consist of 5 different choices marked A to E where one should be the "answer_creator". All data should be in {LANGUAGE}, return only your answer in JSON data format, and add the concept of the data as "question_concepts".

JSON Data:
\end{lstlisting}
\end{prompt}

\subsection{Choosing Model for Data Generation: Indonesian LLM vs. GPT-4 Turbo}
\label{appendix:model_gen_compare}
\begin{table}[t]
\small
\centering
\begin{tabular}{@{}lrrr@{}}
\toprule
\multicolumn{1}{c}{\multirow{2}{*}{\textbf{Model}}} & \multicolumn{3}{c}{\textbf{Win Rate}}                                                                               \\ \cmidrule(l){2-4} 
\multicolumn{1}{c}{}                                & \multicolumn{1}{c}{\textbf{Concept}} & \multicolumn{1}{c}{\textbf{Question}} & \multicolumn{1}{c}{\textbf{Choices}} \\ \midrule
\multicolumn{4}{c}{\textit{Automatic Data Adaptation}}                           \\ \midrule
Merak-v4    & 28.0\%          & 8.5\%           & 4.0\%           \\
GPT-4 Turbo & \textbf{72.0\%} & \textbf{91.5\%} & \textbf{96.0\%} \\ \midrule
\multicolumn{4}{c}{\textit{Automatic Data Generation}}                             \\ \midrule
Merak-v4    & -               & 10.5\%          & 9.5\%           \\
GPT-4 Turbo & -               & \textbf{89.5\%} & \textbf{90.5\%} \\ \bottomrule
\end{tabular}
\caption{Win rate comparison of Merak-v4 (open Indonesian LLM) and GPT-4 Turbo (best-performing proprietary LLM).}
\label{tab:model_gen_compare}
\end{table}
Table \ref{tab:model_gen_compare} shows the win rate accuracy of concept, question, and options generation of Indonesian LLM, Merak-v4, and best-performing proprietary LLM, GPT-4 Turbo. The results demonstrate that GPT-4 Turbo significantly outperforms Merak-v4, with win rates ranging from 72\% to 96\%.
Our sample-level analysis indicates that a significant number of questions generated by Merak-v4 tend to have obvious answers or are formulated as yes/no questions, such as \textit{"Apakah jagung dapat dimakan?"} (\texttt{eng}: Is corn edible?). Additionally, despite being trained on Indonesian texts, Merak-v4 occasionally produces questions with US-centric knowledge. For instance, it generates \textit{"Siapa yang memimpin kampanye pemilihan presiden pertama di Amerika Serikat?"} (\texttt{eng}: Who led the first presidential election campaign in the United States?).

\subsection{Additional Analysis of the LLM-generated Data: Common Mistakes in Concept Adaptation}
\begin{table}[t]
\centering
\small
\begin{tabular}{@{}lrr@{}}
\toprule
\multicolumn{1}{c}{\multirow{2}{*}{\textbf{Error Type}}} & \multicolumn{2}{c}{\textbf{Num (\%) of errors}}                                    \\ \cmidrule(l){2-3} 
\multicolumn{1}{c}{}                                       & \multicolumn{1}{c}{\textbf{\texttt{eng} $\rightarrow$ \texttt{ind}}} & \multicolumn{1}{c}{\textbf{\texttt{ind} $\rightarrow$ \texttt{sun}}} \\ \midrule
Translation                                          & \textbf{4 (57.14\%)}                               & \textbf{31 (86.11\%)}                              \\
Phrase structure                                   & 1 (14.29\%)                               & 1 (2.78\%)                               \\
Typo/mechanics                                    & 2 (28.57\%)                               & 4 (11.11\%)                               \\ \bottomrule
\end{tabular}
\caption{Summary of errors in concept adaptation.}
\label{tab:concept_errors_ind_sun}
\end{table}
The details of common mistakes in concept adaptation for Indonesian and Sundanese are shown in Table \ref{tab:concept_errors_ind_sun}. In the case of adapting concepts from \texttt{eng} to \texttt{ind}, many of the errors are translation errors, resulting from awkward phrasing of the translated concept or the concept remaining in English instead of being translated to Indonesian. However, the number of errors is relatively small. In the case of adapting concepts from \texttt{ind} to \texttt{sun}, similar to \texttt{eng} to \texttt{ind}, the majority of errors also arise from translation errors, with a larger number of errors.

\section{Details on Categories and Concepts}
\label{appendix:categories_and_concepts}
\paragraph{Categories}
Our \textsc{LLM\_Gen} and \textsc{Human\_Gen} datasets include five selected categories, detailed as follows:

\begin{enumerate}
    \item \textbf{Culinary}: Concepts in this category include everything related to culinary, starting from culinary types, cooking ingredients, cooking tools \& methods, to etiquette or eating habits.
    \item \textbf{Places}: Concepts in this category include everything related to places, starting from public facilities, landmarks, buildings, and various other concepts related to places.
    \item \textbf{Culture}: Concepts in this category include everything related to culture, starting from cultural elements, cultural tools, cultural actors, to customs and habits that exist in Indonesia.
    \item \textbf{History}: Concepts in this category include everything related to history, starting from historical events, historical actors, historical findings, and various other concepts related to history.
    \item \textbf{Activities}: Concepts in this category include everything related to activities, starting from sports, hobbies, household work, and various other concepts related to activities.
\end{enumerate}

\paragraph{Concepts Creation}
To create the concepts, we had a human annotator manually list concepts in the selected categories, ensuring they were culturally relevant but general enough for both Indonesian and Sundanese. For example, in the culinary category, concepts might include `duck' or `lime leaves,' avoiding specific dish names. We reviewed the concepts and provided feedback for revisions to ensure quality. This process results in 150 unique concepts per category in Indonesian. The annotators then manually translated these concepts into Sundanese during the QA pairs creation phase (\S\ref{sec:human_gen}). We aimed for parallel concepts to ensure a fair comparison between the languages.

\section{Annotators Demographics}
\label{appendix:anno_demographic}
\begin{table}[t]
\resizebox{\columnwidth}{!}{%
\begin{tabular}{@{}lllc@{}}
\toprule
\multicolumn{1}{c}{\textbf{Data Team}} & \multicolumn{1}{c}{\textbf{Ethnicity}} & \multicolumn{1}{c}{\textbf{Domicile}} & \textbf{Num} \\ \midrule
\multirow{6}{*}{Indonesian}               & \multirow{2}{*}{Sundanese}             & Sukabumi, West Java                   & 1              \\
                           &                            & Depok, West Java       & 1           \\
                           & \multirow{2}{*}{Javanese}  & Magelang, Central Java & 1           \\
                           &                            & Bojonegoro, East Java  & 1           \\
                           & Betawi                     & Tangerang, Banten      & 1           \\
                           & Balinese                   & Denpasar, Bali         & 1           \\ \midrule
\multirow{4}{*}{Sundanese} & \multirow{4}{*}{Sundanese} & Bandung, West Java     & 2           \\
                           &                            & Bogor, West Java       & 1           \\
                           &                            & Majalengka, West Java  & 1           \\
                           &                            & Sukabumi, West Java    & 2           \\ \midrule
\multicolumn{3}{c}{\textbf{Total}}                                               & \textbf{12} \\ \bottomrule
\end{tabular}%
}
\caption{Demographic information of the annotators from each dataset team. Note that even though Depok is included in West Java Province and Tangerang is included in Banten Province, both are geographically closer to Jakarta and considered as part of the Greater Jakarta area (\textit{Jabodetabek}).}
\label{tab:anno_demographic}
\end{table}
In accordance with the aim of constructing the dataset, we involved 6 Indonesian and 6 Sundanese native-speaker annotators. We chose annotators who have prior experience in building NLP datasets in Indonesian and/or Sundanese through close recruitment. All the annotators who worked on the Indonesian dataset were people from several regions on the islands of Java and Bali. Meanwhile, the annotators involved in building the Sundanese dataset were Sundanese people who come from several different regions in West Java. The detailed annotators' demographics are shown in Table \ref{tab:anno_demographic}.

\section{Human Annotation Guideline}
\label{appendix:anno_guideline}
To ensure a high-quality and standardized format for question-answers annotation, we provide a specific guideline during the annotation process. The process of creating question-answer pair data is carried out using Google Spreadsheets. Each person will get a Google Sheets document that will be their worksheet. The explanation of each field is described below.

\begin{enumerate}
    \item \textbf{ID}: This column contains the ID of each data.
    \item \textbf{Category}: This column contains the overarching category of the data.
    \item \textbf{Question Concept}: This column contains concepts from categories that need to be used in creating questions.
    \item \textbf{Question}: This column is used to write commonsense questions that contain the given concept and match the category.
    \item \textbf{Choices}: This column is used to write 5 choices for the questions given. Of the 5 choices given, the annotator needs to ensure there is 1 correct answer and 1 distractor. Distractor is an answer that could potentially be considered the correct answer.
    \item \textbf{Answer}: This column only needs to be filled in when the row contains the correct answer.
    \item \textbf{Distractor}: This column only needs to be filled in when the row contains the distractor.
\end{enumerate}

\paragraph{General Rules}
There are several rules in the data creation process, detailed as follows:
\begin{enumerate}
    \item The questions asked are commonsense questions (not factual questions) related to Indonesian/Sundanese culture. Especially for the History category, questions can also be in the form of factual questions, but they must be general facts (commonly known).
    \item The questions created must be related to the given category and must contain the given concept.
    \item Annotators are allowed to change the morphological form of concepts as long as they do not change the categories and basic words. (i.e. \textit{ber\textbf{kunjung}} $\rightarrow$ \textit{\textbf{kunjung}an}, \textit{meng\textbf{unjung}i}, \textit{di\textbf{kunjung}i}).
    \item The priority is that the questions asked are always related to general daily life or Indonesian/Sundanese culture.
    \item Both language groups will use the same list of categories and concepts. However, when creating data, annotators are expected to incorporate perspectives that align with their respective cultural backgrounds.
    \item Each person will get 50 concepts from each category. The total number of concepts that will be accepted is 250 concepts, equivalent to the workload for each person, where the expectation is to create one question per concept.
\end{enumerate} 

\section{Manual Data Generation Findings}
\label{appendix:manual_data_findings}
During the manual data generation process, we discovered several interesting findings from the data created by the annotators. Note that the common errors within these findings occur before the quality control meeting (\S\ref{sec:human_gen}). In our final dataset, we have ensured that such problems have been fixed.

\paragraph{Questions}
We found quite a lot of interesting findings in the process of creating commonsense questions. We found that there were questions created using very general contexts, while we required the data generation to be scoped within the Indonesian and Sundanese contexts (specifically for the Sundanese dataset). For instance, an annotator produced \textit{"Apa jenis restoran yang paling terkenal di seluruh dunia?"} (\texttt{eng}: Which type of restaurant is most famous worldwide?) In that question, the annotator developed the concept question \textit{“restoran”} (\texttt{eng}: restaurant) into a question with a very general context: \textit{“di seluruh dunia"} (\texttt{eng}: all over the world). We also found subjective questions, so the answers that emerged also had subjective value (not commonsense). These subjective questions are usually characterized by the use of superlative adjectives. Another type of finding related to the question category is the creation of logical/causal questions. This seems to be based on confusion from annotators regarding the boundaries of commonsense and logic/causation.

\paragraph{Options}
The findings related to options that are most often encountered are making choices that have the same value as each other so that it is difficult to determine the answer and distractor. For instance, the question \textit{"Kue apa yang biasanya disajikan pada momen lebaran?"} (\texttt{eng}: What cakes are usually served during Eid?). All the options given are types of cakes that are generally served during Eid. This finding is also related to the issue of subjectivity.

\paragraph{Answers-Distractors}
Answers and distractors are another category that also has many findings in the manual data generation process. This issue is not much different from the findings in Options: answers and distractors are equivalent. Apart from that, in this category, it was also found that personal experience was used in determining answers so that more common answers were determined as distractors. For instance, the question \textit{"Apa yang biasanya orang lakukan di stasiun?"} (\texttt{eng}: What do people usually do at the train station?). The annotator specified \textit{"mengantar teman/kerabat"} (\texttt{eng}: accompany friends/relatives) as the answer. While the \textit{"naik-turun kereta"} (\texttt{eng}: get on and off the train) option was chosen as a distractor.

\paragraph{Knowledge Variations between Annotators}
\begin{figure}[t!]
    \begin{center}
        \centerline{\includegraphics[width=\columnwidth]{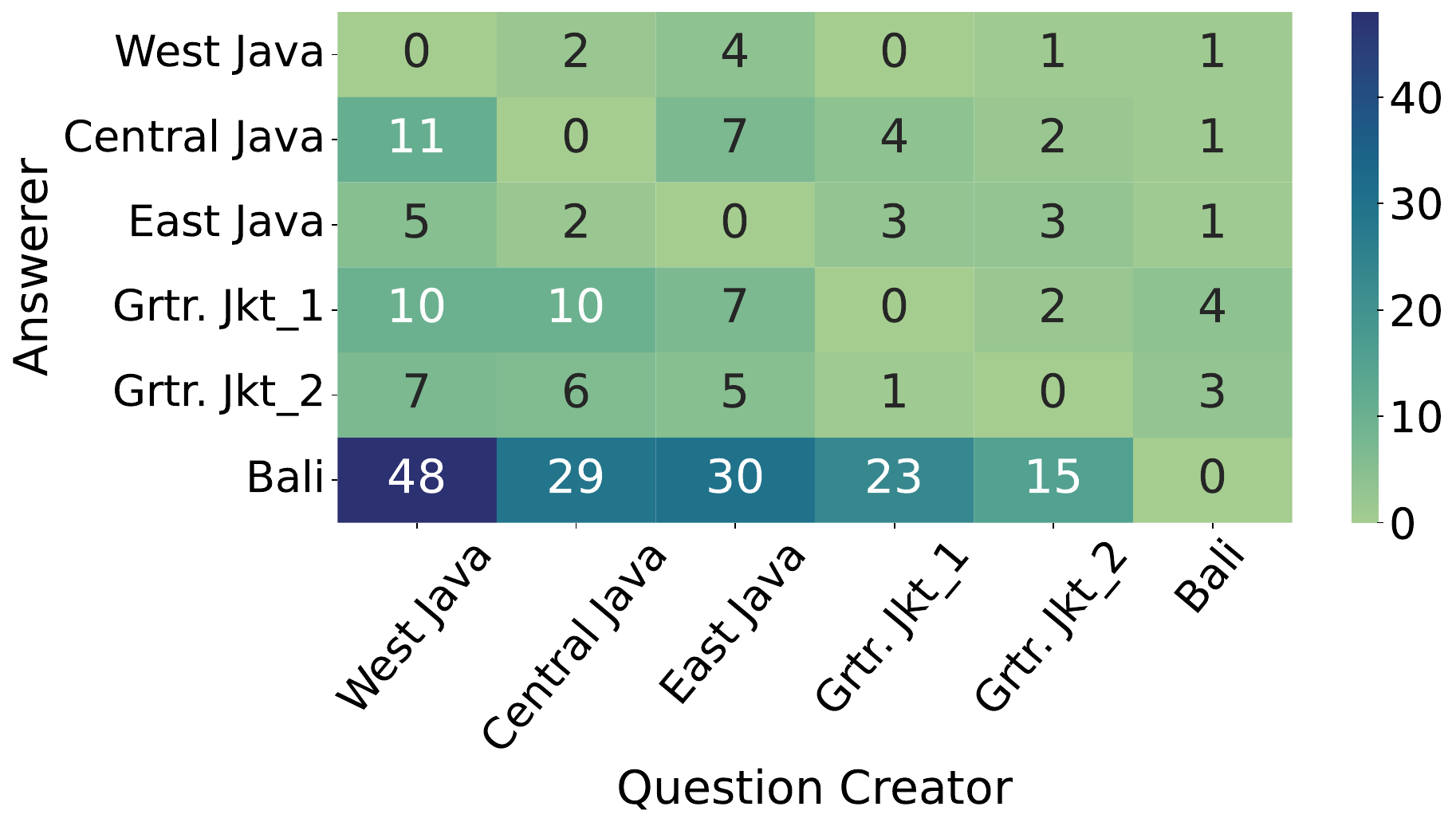}}
        \caption{Answer conflict across Indonesian annotators.}
        \label{fig:anno_conlict_ind}
    \end{center}
\end{figure}

\begin{figure}[t!]
    \begin{center}
        \centerline{\includegraphics[width=\columnwidth]{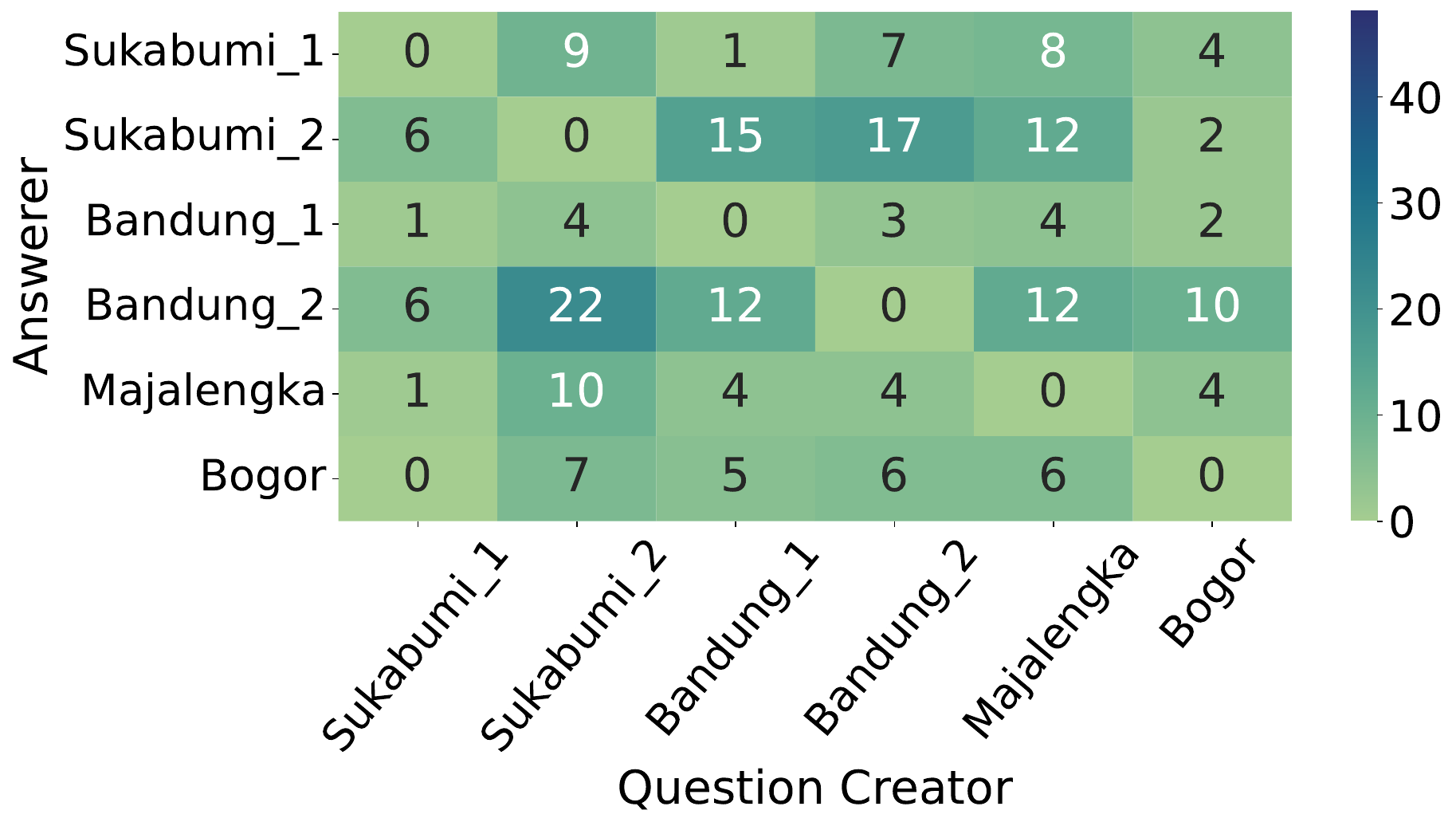}}
        \caption{Answer conflict across Sundanese annotators.}
        \label{fig:anno_conlict_sun}
    \end{center}
\end{figure}

As we employ annotators from different regions, we can analyze variations in the data they generate. We examine this by calculating the number of answer conflicts that arise during the "answering question" phase of our data generation pipeline (\S\ref{sec:human_gen}). The results from Indonesian and Sundanese annotators are detailed in Figure \ref{fig:anno_conlict_ind} and \ref{fig:anno_conlict_sun}, respectively.

Our analysis revealed that, out of the Indonesian annotators, the one from Bali has the highest number of conflicting answers. However, the number of questions generated by the Bali annotator does not seem to have a lot of conflicts, suggesting that this annotator tends to generate easier questions, some of which have an obvious answer. This finding contrasts with the results from West Java annotators, who, despite generating questions that lead to a higher number of conflicts, do so mainly due to the creation of more challenging option distractors.

Interestingly, in the case of Sundanese annotators, the variation in answer conflicts across regions is not significant. This lack of variation can likely be attributed to the Sundanese language's narrower geographic distribution which primarily spoken in West Java, unlike the Indonesian language, which serves as a lingua franca within the country. However, it is important to note that given the small number of annotators, these trends cannot be conclusively linked to the annotators' regional characteristics. The observed differences may also be due to individual differences rather than regional ones.

\section{Zero-Shot Benchmark Prompts}
\label{appendix:benchmark_prompt}
We apply three prompt variations to test LLMs' performance on our CommonsenseQA datasets. Each prompt is described below.

\begin{flushleft}
    \textbf{Prompt Variation 1}
\end{flushleft}
\begin{prompt}
\begin{lstlisting}
The following are multiple choice questions (with answers) about "{CONCEPT}".
{QUESTION}
A. {CHOICE_A}
B. {CHOICE_B}
C. {CHOICE_C}
D. {CHOICE_D}
E. {CHOICE_E}
Answer:
\end{lstlisting}
\end{prompt}

\begin{flushleft}
    \textbf{Prompt Variation 2}
\end{flushleft}
\begin{prompt}
\begin{lstlisting}
Question: {QUESTION}
Choices:
A. {CHOICE_A}
B. {CHOICE_B}
C. {CHOICE_C}
D. {CHOICE_D}
E. {CHOICE_E}
Answer:
\end{lstlisting}
\end{prompt}

\begin{flushleft}
    \textbf{Prompt Variation 3}
\end{flushleft}
\begin{prompt}
\begin{lstlisting}
The following are multiple choice questions (with answers) about "{CONCEPT}".
Question: {QUESTION}
A. {CHOICE_A}
B. {CHOICE_B}
C. {CHOICE_C}
D. {CHOICE_D}
E. {CHOICE_E}
Answer:
\end{lstlisting}
\end{prompt}

\begin{figure}[t!]
    \begin{center}
        \centerline{\includegraphics[width=\columnwidth]{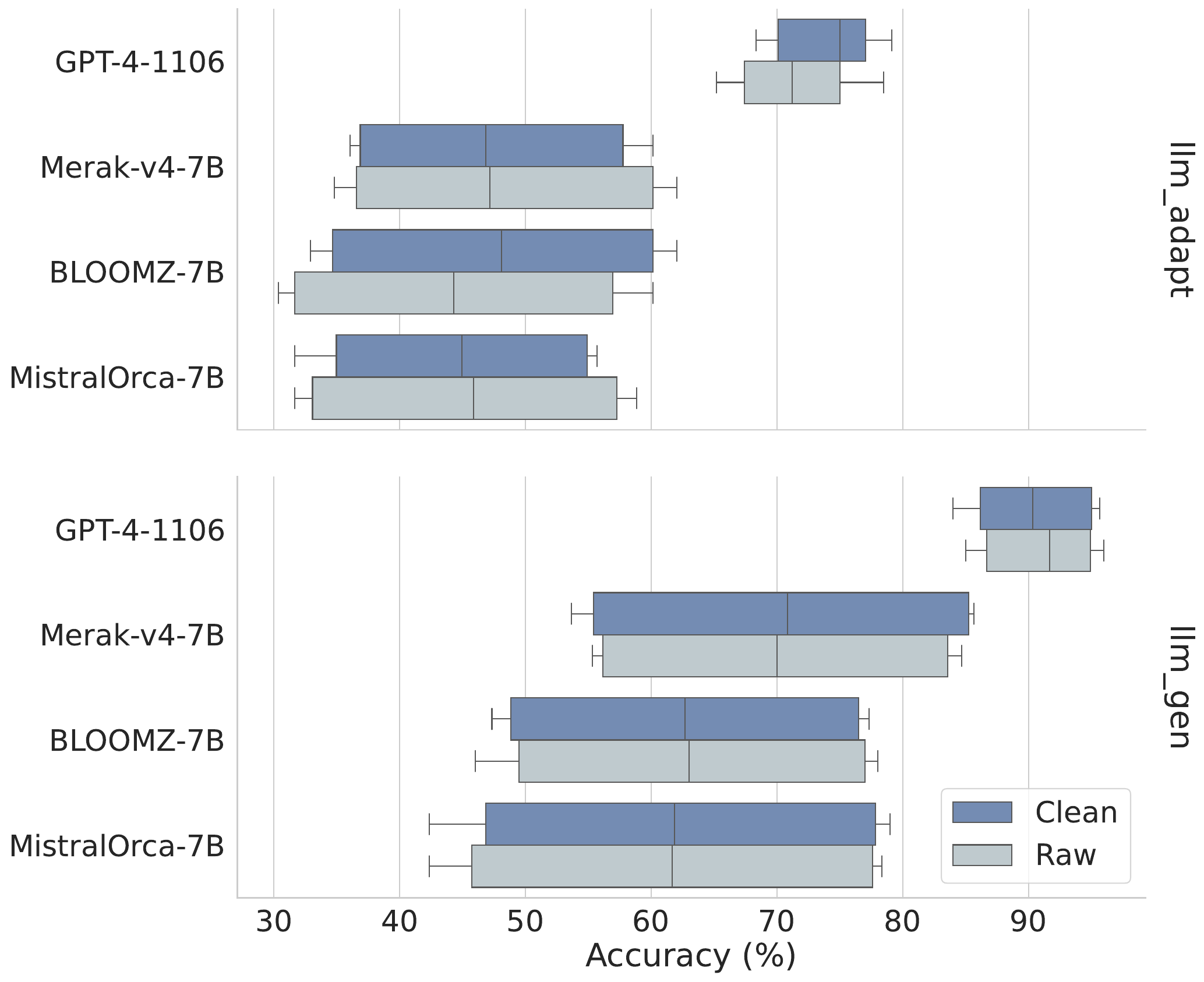}}
        \caption{Performance comparison of raw vs. cleaned version of LLM-generated data.}
        \label{fig:raw_clean_gap}
    \end{center}
\end{figure}

\section{Effect of Synthetic Dataset Cleaning}
\label{appendix:data_cleaning}
To check how much cleaning the noise in synthetic (LLM-generated) data affects LLMs performance in answering our questions, we manually correct the errors in all 158 test sets of \textsc{LLM\_Adapt} and 300 randomly sampled data from the \textsc{LLM\_Gen}. The results are presented in Figure \ref{fig:raw_clean_gap}. The performance variance appears smaller on the cleaned dataset for \textsc{LLM\_Adapt}, leading to more consistent performance. However, the performance improvement is not as significant for the \textsc{LLM\_Gen} dataset, likely due to the lesser noise in this dataset compared to \textsc{LLM\_Adapt}. Thus, a robust model capable of producing cleaner data can minimize the need for extensive dataset cleaning. However, caution is still necessary as errors can still occur.

\begin{table*}[t]
\resizebox{\textwidth}{!}{%
\begin{tabular}{@{}lll@{}}
\toprule
\textbf{Classification} &
  \textbf{Original Data (\texttt{eng})} &
  \textbf{LLM Adapted Data (\texttt{ind})} \\ \midrule
\begin{tabular}[c]{@{}l@{}}concept=\red{irrelevant}\\ name=\red{irrelevant}\\ option=\red{irrelevant}\end{tabular} &
  \begin{tabular}[c]{@{}l@{}}Concept: \red{koala}\\ Question: \red{James} looked for \red{koalas}, but\\ misunderstood and went to the wrong environment\\ where is it impossible to find \red{koalas}?\\ Options:\\ A. jungle\\ B. great outdoors\\ C. \red{siberia}\\ D. \red{queensland}\\ E. wilderness\end{tabular} &
  \begin{tabular}[c]{@{}l@{}}Concept: \green{orangutan}\\ Question: \green{Budi} mencari \green{orangutan}, tetapi \\ salah paham dan pergi ke lingkungan yang salah.\\ Di mana mustahil untuk menemukan \green{orangutan}?\\ Options: \\ A. hutan hujan tropis\\ B. alam bebas\\ C. \red{gurun sahara}\\ D. \green{kalimantan}\\ E. belantara\end{tabular} \\ \midrule
\begin{tabular}[c]{@{}l@{}}concept=relevant\\ name=\red{irrelevant}\\ option=\red{irrelevant}\end{tabular} &
  \begin{tabular}[c]{@{}l@{}}Concept: hotel room\\ Question: \red{John} rented a hotel room while on a\\ business trip. It was a company sponsored event.\\ Where might it have been?\\ Options:\\ A. convention center\\ B. business interviews\\ C. \red{michigan}\\ D. resort\\ E. city\end{tabular} &
  \begin{tabular}[c]{@{}l@{}}Concept: kamar hotel\\ Question: \green{Andi} menyewa kamar hotel saat dalam\\ perjalanan bisnis. Ini adalah acara yang disponsori\\ perusahaannya. Di mana acara tersebut mungkin\\ diadakan?\\ Options:\\ A. pusat konvensi\\ B. wawancara bisnis\\ C. \green{jakarta}\\ D. resor\\ E. kota\end{tabular} \\ \midrule
\begin{tabular}[c]{@{}l@{}}concept=\red{irrelevant}\\ name=relevant\\ option=\red{irrelevant}\end{tabular} &
  \begin{tabular}[c]{@{}l@{}}Concept: \red{blue ridge mountains}\\ Question: Which state is the \red{blue ridge mountains}\\ \red{roanoke city} in?\\ Options:\\ A. \red{appalachians}\\ B. \red{north carolina}\\ C. \red{virginia}\\ D. \red{pennsylvania}\\ E. \red{montana}\end{tabular} &
  \begin{tabular}[c]{@{}l@{}}Concept: \green{Gunung Bromo}\\ Question: Di provinsi mana \green{Gunung Bromo} terletak?\\ Options:\\ A. \green{jawa barat}\\ B. \green{jawa tengah}\\ C. \green{jawa timur}\\ D. \green{yogyakarta}\\ E. \green{bali}\end{tabular} \\ \midrule
\begin{tabular}[c]{@{}l@{}}concept=relevant\\ name=relevant\\ option=relevant\end{tabular} &
  \begin{tabular}[c]{@{}l@{}}Concept: teacher\\ Question: What does a kindergarten teacher do\\ before nap time?\\ Options:\\ A. lower expectations\\ B. encourage\\ C. fear\\ D. time test\\ E. tell story\end{tabular} &
  N/A \\ \bottomrule
\end{tabular}
}
\caption{Examples of the concept, name, and option relevancy classification result. The English data instance that is classified as irrelevant in any of these three elements will be adapted.}
\label{tab:concept_classification_example}
\end{table*}
\begin{table*}[t]
\small
\centering
\resizebox{\textwidth}{!}{%
\begin{tabular}{@{}lllllc@{}}
\toprule
\textbf{Category} &
  \textbf{Concept} &
  \textbf{Data Creator} &
  \textbf{Question} &
  \textbf{Choices} &
  \textbf{Answer} \\ \midrule
\multirow{2}{*}{activity} &
  \multirow{2}{*}{\begin{tabular}[c]{@{}l@{}}panen \\ (\textit{harvest})\end{tabular}} &
  GPT-4 Turbo &
  \begin{tabular}[c]{@{}l@{}}Kegiatan apa yang biasanya dilakukan \\ ketika musim panen tiba? \\ (\textit{What activity is usually done when} \\ \textit{the harvest season arrives?})\end{tabular} &
  \begin{tabular}[c]{@{}l@{}}A. Makan bersama keluarga \\ (\textit{eating together with family})\\ B. Berlibur (\textit{going on a holiday})\\ C. Memetik hasil pertanian \\ (\textit{picking agricultural produce})\\ D. Memancing (\textit{fishing})\\ E. Berbelanja (\textit{shopping})\end{tabular} &
  C \\ \cmidrule(l){3-6}
 &
   &
  Human &
  \begin{tabular}[c]{@{}l@{}}Kegiatan apa yang masyarakat desa lakukan \\ setelah panen mereka berhasil? \\ (\textit{What activity do village communities do}\\  \textit{after their harvest succeeds}?)\end{tabular} &
  \begin{tabular}[c]{@{}l@{}}A. Bercocok tanam (\textit{farming})\\ B. Memotong rumput \\ (\textit{cutting grass})\\ C. Syukuran \\ (\textit{having a `syukuran' event})\\ D. Jalan-jalan bersama \\ (\textit{going out together})\\ E. Wortel (\textit{carrots})\end{tabular} &
  C \\ \midrule
\multirow{2}{*}{culinary} &
  \multirow{2}{*}{\begin{tabular}[c]{@{}l@{}}kerupuk \\ (\textit{crackers})\end{tabular}} &
  GPT-4 Turbo &
  \begin{tabular}[c]{@{}l@{}}Kerupuk biasanya terbuat dari \\ bahan dasar apa? \\ (\textit{What is the common primary} \\ \textit{ingredient of crackers?})\end{tabular} &
  \begin{tabular}[c]{@{}l@{}}A. Tepung terigu (\textit{wheat flour})\\ B. Tepung beras (\textit{rice flour})\\ C. Tepung tapioka (\textit{tapioca flour})\\ D. Tepung jagung (\textit{corn flour})\\ E. Tepung kentang (\textit{potato flour})\end{tabular} &
  C \\ \cmidrule(l){3-6}
 &
   &
  Human &
  \begin{tabular}[c]{@{}l@{}}Terbuat dari apakah kerupuk rambak? \\ (\textit{What is `rambak' crackers made of?})\end{tabular} &
  \begin{tabular}[c]{@{}l@{}}A. Kulit hewan (\textit{animal skin})\\ B. Lidah hewan (\textit{animal tounge})\\ C. Usus hewan (\textit{animal intestines})\\ D. Nasi (\textit{rice})\\ E. Tepung beras (\textit{rice flour})\end{tabular} &
  A \\ \midrule
\multirow{2}{*}{culture} &
  \multirow{2}{*}{\begin{tabular}[c]{@{}l@{}}tarian\\ (\textit{dance})\end{tabular}} &
  GPT-4 Turbo &
  \begin{tabular}[c]{@{}l@{}}Tarian apa yang merupakan \\ tarian tradisional dari Bali? \\ (\textit{Which dance is a traditional} \\ \textit{dance from Bali}?)\end{tabular} &
  \begin{tabular}[c]{@{}l@{}}A. Tari Pendet (\textit{Pendet dance})\\ B. Tari Saman (\textit{Saman dance})\\ C. Tari Kecak (\textit{Kecak dance})\\ D. Tari Jaipong (\textit{Jaipong dance})\\ E. Tari Yapong (\textit{Yapong dance})\end{tabular} &
  C \\ \cmidrule(l){3-6}
 &
   &
  Human &
  \begin{tabular}[c]{@{}l@{}}Manakah yang termasuk tarian dari Bali? \\ (\textit{Which includes a dance from Bali?})\end{tabular} &
  \begin{tabular}[c]{@{}l@{}}A. Tari Piring (\textit{Piring Dance})\\ B. Tari Pendet (\textit{Pendet Dance})\\ C. Tari Topeng (\textit{Topeng Dance})\\ D. Tari Sajojo (\textit{Sajojo Dance})\\ E. Tari Melasti (\textit{Melasti Dance})\end{tabular} &
  B \\ \midrule
\multirow{2}{*}{history} &
  \multirow{2}{*}{\begin{tabular}[c]{@{}l@{}}era\\ (\textit{era})\end{tabular}} &
  GPT-4 Turbo &
  \begin{tabular}[c]{@{}l@{}}Era apa yang ditandai dengan berakhirnya \\ Perang Dingin dan runtuhnya Uni Soviet? \\ (\textit{Which era is marked by the end of the} \\ \textit{Cold War and the collapse of} \\ \textit{the Soviet Union}?)\end{tabular} &
  \begin{tabular}[c]{@{}l@{}}A. Era Globalisasi \\ (\textit{globalization era})\\ B. Era Informasi (\textit{information era})\\ C. Era Reformasi (\textit{reformation era})\\ D. Era Pasca-Perang Dingin \\ (\textit{pasca-cold war era})\\ E. Era Industrialisasi \\ (\textit{industrialization era})\end{tabular} &
  D \\ \cmidrule(l){3-6}
 &
   &
  Human &
  \begin{tabular}[c]{@{}l@{}}Berapa lama era Orde Baru \\ berlangsung di Indonesia? \\ (\textit{How long did the New Order era} \\ \textit{last in Indonesia}?)\end{tabular} &
  \begin{tabular}[c]{@{}l@{}}A. 4 tahun (\textit{4 years})\\ B. 5 tahun (\textit{5 years})\\ C. 20 tahun (\textit{20 years})\\ D. 25 tahun (\textit{25 years})\\ E. 32 tahun (\textit{32 years})\end{tabular} &
  E \\ \midrule
\multirow{2}{*}{place} &
  \multirow{2}{*}{\begin{tabular}[c]{@{}l@{}}gunung \\ (\textit{mountain})\end{tabular}} &
  GPT-4 Turbo &
  \begin{tabular}[c]{@{}l@{}}Di manakah biasanya gunung berada? \\ (\textit{Where are mountains usually located?})\end{tabular} &
  \begin{tabular}[c]{@{}l@{}}A. Di kota (\textit{in the city})\\ B. Di dataran tinggi (\textit{in highlands})\\ C. Di dalam laut (\textit{in the sea})\\ D. Di padang pasir (\textit{in the desert})\\ E. Di pinggir sungai \\ (\textit{at the riverbank})\end{tabular} &
  B \\ \cmidrule(l){3-6}
 &
   &
  Human &
  \begin{tabular}[c]{@{}l@{}}Apa nama gunung tertinggi \\ di Indonesia? \\ (\textit{What is the name of the} \\ \textit{highest mountain in Indonesia?})\end{tabular} &
  \begin{tabular}[c]{@{}l@{}}A. Gunung Jaya Wijaya \\ (\textit{Mt. Jaya Wijaya})\\ B. Gunung Bromo (\textit{Mt. Bromo})\\ C. Gunung Merapi (\textit{Mt. Merapi})\\ D. Gunung Semeru (\textit{Mt. Semeru})\\ E. Gunung Sumbing (\textit{Mt. Sumbing})\end{tabular} &
  A \\ \bottomrule
\end{tabular}%
}
\caption{Examples of data generated by GPT-4 Turbo (\texttt{gpt-4-1106-preview}) compared to human annotators.}
\label{tab:llm_vs_human_gen_examples}
\end{table*}
\section{LLM vs. Human-Generated Data Examples}
We provide some GPT-4 Turbo generation samples compared with human-generated data in Table \ref{tab:llm_vs_human_gen_examples}.

\begin{table*}[htbp]
\small
\centering
\begin{tabular}{|p{0.31\textwidth}|p{0.31\textwidth}|p{0.31\textwidth}|}
\toprule
\textbf{Prompt} & \textbf{Response} & \textbf{Failure Reason} \\ \midrule
Apa judul lagu yang wajib dinyanyikan pada sesi mengheningkan cipta di setiap upacara bendera? \newline \textit{(What is the mandatory song to be sung during the `mengheningkan cipta' (moment of silence) in every flag ceremony?)} & Pada sesi mengheningkan cipta di setiap upacara bendera di Indonesia, biasanya \red{tidak dinyanyikan lagu}. ... \newline \textit{(During the moment of silence in every flag ceremony in Indonesia, \red{a song is usually not sung}. ...)} & The mandatory song to be sung during the \textit{`mengheningkan cipta'} (moment of silence) session in flag ceremonies in Indonesia is \textit{`Hymne Pahlawan (Mengheningkan Cipta).'} \\ \midrule
Manakah menu yang dikonsumsi menggunakan sumpit? \newline \textit{(Which dishes are consumed with chopsticks?)} & Sumpit adalah alat makan yang biasanya digunakan dalam tradisi kuliner Asia Timur, seperti Jepang, Korea, dan China. Berikut adalah beberapa contoh menu yang sering dikonsumsi menggunakan sumpit: \red{Sushi dan Sashimi (Jepang)}, \red{Ramen (Jepang)}, ... \textit{\red{[proceed with other dish examples in Japan, Korea, and China]}} \newline \textit{(Chopsticks are utensils that are commonly used in East Asian culinary traditions, such as Japan, Korea, and China. Here are some examples of dishes often consumed with chopsticks: \red{Sushi and Sashimi (Japan)}, \red{Ramen (Japan)}, ... \red{[proceed with other dish examples in Japan, Korea, and China]})} & Many Indonesian foods are commonly eaten with chopsticks, such as \textit{mie ayam} (chicken noodles), \textit{bakmi}, \textit{mie yamin}, and many more. \\ \midrule
Di daerah mana yang terdapat banyak vila? \newline \textit{(In which area can we find many villas?)} & ... beberapa daerah yang terkenal dengan banyaknya vila adalah: 1. Bali, Indonesia 2. \red{Tuscany, Italia} 3. \red{Provence, Prancis} \textit{\red{[proceed with other location in Spain, US, Thailand, Portugal, US]}} \newline \textit{(... some areas known for having many villas include: 1. Bali, Indonesia 2. \red{Tuscany, Italy} 3. \red{Provence, France} \red{[proceed with other locations in Spain, US, Thailand, Portugal, US]})} & Apart from Bali, there are many other areas in Indonesia famous for having many villas, such as Bandung, Lembang, Puncak, Batu, and many more. \\ \bottomrule
\end{tabular}
\caption{Examples of incorrect responses by GPT-4 Turbo (\texttt{gpt-4-1106-preview}) in `free' or open-ended generation settings.}
\label{tab:open_gen_examples}
\end{table*}
\section{Examples from Multiple-Choice vs. `Free' Generation Experiment}
We provide some GPT-4 Turbo generation samples when given the question in `free' or open-ended generation settings in Table \ref{tab:open_gen_examples}.

\end{document}